\def\year{2021}\relax
\documentclass[letterpaper]{article} 
\usepackage{aaai21}  
\usepackage{times}  
\usepackage{helvet} 
\usepackage{courier}  
\usepackage[hyphens]{url}  
\usepackage{graphicx} 
\usepackage{natbib}  
\usepackage{caption} 
\usepackage{times}
\usepackage{soul}
\usepackage{url}
\usepackage{graphicx}
\usepackage{amsmath}
\usepackage{amsthm}
\usepackage{amsfonts}
\usepackage{algorithm}
\usepackage{algorithmic}
\usepackage{caption}
\usepackage{multirow}
\usepackage{booktabs}
\usepackage{threeparttable}
\usepackage{enumitem}
\usepackage{footmisc}
\usepackage{subfigure}
\usepackage{multibib}
\newcites{apndx}{References in Appendix}
\newtheorem{lemma}{\textbf{Lemma}}
\newtheorem{proposition}{\textbf{Proposition}}
\newcommand{\mb}{\mathbf}
\newcommand{\bigO}{\mathcal{O}}

\newcommand{\mn}{Informer}

\nocopyright

\usepackage[toc,title,page]{appendix}

\title{Informer: Beyond Efficient Transformer for Long Sequence \\ Time-Series Forecasting}
\author{
    Haoyi Zhou, \textsuperscript{\rm 1}
    Shanghang Zhang, \textsuperscript{\rm 2}
    Jieqi Peng, \textsuperscript{\rm 1}
    Shuai Zhang, \textsuperscript{\rm 1}
    Jianxin Li, \textsuperscript{\rm 1} \\
    Hui Xiong, \textsuperscript{\rm 3}
    Wancai Zhang \textsuperscript{\rm 4} \\
}
\affiliations{
    \textsuperscript{\rm 1} Beihang University
    \textsuperscript{\rm 2} UC Berkeley
    \textsuperscript{\rm 3} Rutgers University
    \textsuperscript{\rm 4} SEDD Company \\
    \{zhouhy, pengjq, zhangs, lijx\}@act.buaa.edu.cn, shz@eecs.berkeley.edu, \{xionghui,zhangwancaibuaa\}@gmail.com \\
    }

\begin{document}

\maketitle

\begin{abstract}

Many real-world applications require the prediction of long sequence time-series, such as electricity consumption planning. Long sequence time-series forecasting (LSTF) demands a high prediction capacity of the model, which is the ability to capture precise long-range dependency coupling between output and input efficiently. Recent studies have shown the potential of Transformer to increase the prediction capacity. However, there are several severe issues with Transformer that prevent it from being directly applicable to LSTF, including quadratic time complexity, high memory usage, and inherent limitation of the encoder-decoder architecture. 
To address these issues, we design an efficient transformer-based model for LSTF, named Informer, with three distinctive characteristics: (i) a \emph{ProbSparse} self-attention mechanism, which achieves $\bigO(L \log L)$ in time complexity and memory usage, and has comparable performance on sequences' dependency alignment. (ii) the self-attention distilling highlights dominating attention by halving cascading layer input, and efficiently handles extreme long input sequences. (iii) the generative style decoder, while conceptually simple, predicts the long time-series sequences at one forward operation rather than a step-by-step way, which drastically improves the inference speed of long-sequence predictions. Extensive experiments on four large-scale datasets demonstrate that Informer significantly outperforms existing methods and provides a new solution to the LSTF problem.

\end{abstract}

\section{Introduction}

\begin{figure}[t]
\centering
\subfigure[Sequence Forecasting.]{
\includegraphics[width=0.46\linewidth]{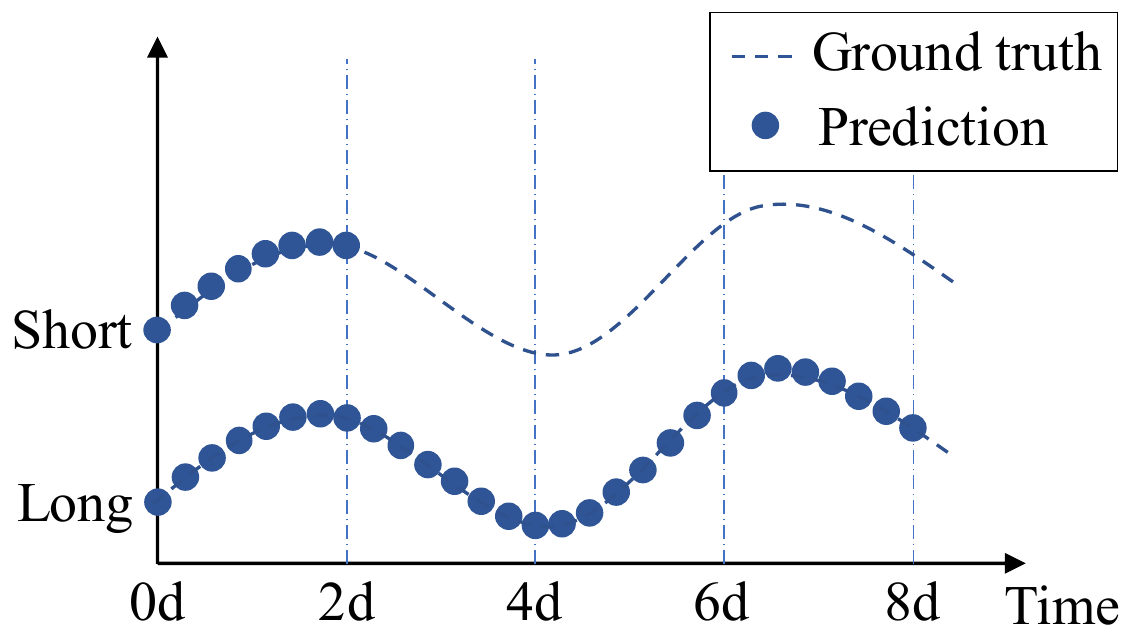}
}
\hfill
\subfigure[Run LSTM on sequences.]{
\includegraphics[width=0.48\linewidth]{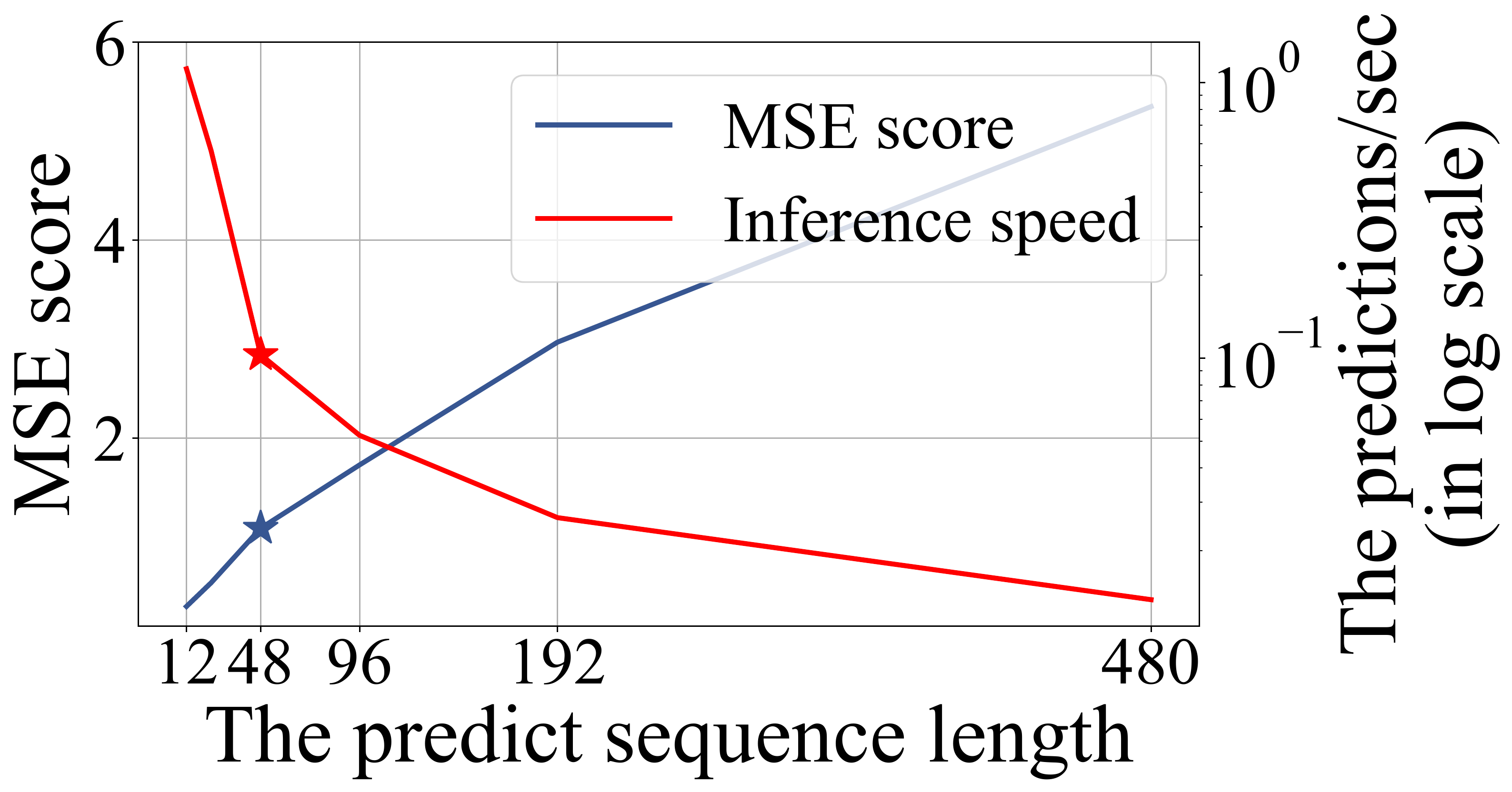}
}
\caption{(a) LSTF can cover an extended period than the short sequence predictions, making vital distinction in policy-planning and investment-protecting. (b) The prediction capacity of existing methods limits LSTF's performance. E.g., starting from length=48, MSE rises unacceptably high, and the inference speed drops rapidly.}
\label{fig:intro.4fig}
\end{figure}

\noindent Time-series forecasting is a critical ingredient across many domains, such as sensor network monitoring \cite{papadimitriou2006optimal}, energy and smart grid management, economics and finance \cite{DBLP:conf/vldb/ZhuS02}, and disease propagation analysis \cite{DBLP:conf/kdd/MatsubaraSPF14}. In these scenarios, we can leverage a substantial amount of time-series data on past behavior to make a forecast in the long run, namely long sequence time-series forecasting (LSTF).
However, existing methods are mostly designed under short-term problem setting, like predicting 48 points or less \cite{hochreiter1997long, li2017graph, yu2017long, journals/corr/abs-1904-07464, conf/ijcai/QinSCCJC17,wen2017multi}. 
The increasingly long sequences strain the models' prediction capacity to the point where this trend is holding the research on LSTF.
As an empirical example, Fig.(\ref{fig:intro.4fig}) shows the forecasting results on a real dataset, where the LSTM network predicts the hourly temperature of an electrical transformer station from the short-term period (12 points, 0.5 days) to the long-term period (480 points, 20 days). The overall performance gap is substantial when the prediction length is greater than 48 points (the solid star in Fig.(\ref{fig:intro.4fig}b)), where the MSE rises to unsatisfactory performance, the inference speed gets sharp drop, and the LSTM model starts to fail.

The major challenge for LSTF is to enhance the prediction capacity to meet the increasingly long sequence demand, which requires (a) extraordinary long-range alignment ability and (b) efficient operations on long sequence inputs and outputs.
Recently, Transformer models have shown superior performance in capturing long-range dependency than RNN models. The self-attention mechanism can reduce the maximum length of network signals traveling paths into the theoretical shortest $\bigO(1)$ and avoid the recurrent structure, whereby Transformer shows great potential for the LSTF problem.
Nevertheless, the self-attention mechanism violates requirement (b) due to its $L$-quadratic computation and memory consumption on $L$-length inputs/outputs. Some large-scale Transformer models pour resources and yield impressive results on NLP tasks \cite{DBLP:journals/corr/abs-2005-14165}, but the training on dozens of GPUs and expensive deploying cost make theses models unaffordable on real-world LSTF problem. The efficiency of the self-attention mechanism and Transformer architecture becomes the bottleneck of applying them to LSTF problems.
Thus, in this paper, we seek to answer the question: \emph{can we improve Transformer models to be computation, memory, and architecture efficient, as well as maintaining higher prediction capacity?}

Vanilla Transformer~\cite{vaswani2017attention} has three significant limitations when solving the LSTF problem:

\begin{enumerate}[nosep, leftmargin=0.45cm]
    \item \textbf{The quadratic computation of self-attention.} The atom operation of self-attention mechanism, namely canonical dot-product, causes the time complexity and memory usage per layer to be $\bigO(L^2)$. 
    \item \textbf{The memory bottleneck in stacking layers for long inputs.} The stack of $J$ encoder/decoder layers makes total memory usage to be $\bigO(J \cdot L^2)$, which limits the model scalability in receiving long sequence inputs.
    \item \textbf{The speed plunge in predicting long outputs.} Dynamic decoding of vanilla Transformer makes the step-by-step inference as slow as RNN-based model (Fig.(\ref{fig:intro.4fig}b)).
\end{enumerate}
There are some prior works on improving the efficiency of self-attention. The Sparse Transformer \cite{child2019generating}, LogSparse Transformer \cite{li2019enhancing}, and Longformer \cite{DBLP:journals/corr/abs-2004-05150} all use a heuristic method to tackle limitation 1 and reduce the complexity of self-attention mechanism to $\bigO(L \log L)$, where their efficiency gain is limited \cite{qiu2019blockwise}. Reformer \cite{kitaev2019reformer} also achieves $\bigO(L \log L)$ with locally-sensitive hashing self-attention, but it only works on extremely long sequences. More recently, Linformer \cite{wang2020linformer} claims a linear complexity $\bigO(L)$, but the project matrix can not be fixed for real-world long sequence input, which may have the risk of degradation to $\bigO(L^2)$. Transformer-XL \cite{dai2019transformer} and Compressive Transformer \cite{rae2019compressive} use auxiliary hidden states to capture long-range dependency, which could amplify limitation 1 and be adverse to break the efficiency bottleneck. All these works mainly focus on limitation 1, and the limitation 2\&3 remains unsolved in the LSTF problem. To enhance the prediction capacity, we tackle all these limitations and achieve improvement beyond efficiency in the proposed \mn.

To this end, our work delves explicitly into these three issues.  We investigate the sparsity in the self-attention mechanism, make improvements of network components, and conduct extensive experiments.
The contributions of this paper are summarized as follows:

\begin{itemize}[nosep, leftmargin=0.45cm]
\item We propose {\mn} to successfully enhance the prediction capacity in the LSTF problem, which validates the Transformer-like model's potential value to capture individual long-range dependency between long sequence time-series outputs and inputs.
\item We propose \emph{ProbSparse} self-attention mechanism to efficiently replace the canonical self-attention. It achieves the $\bigO(L \log L)$ time complexity and $\bigO(L \log L)$ memory usage on dependency alignments.
\item We propose self-attention distilling operation to privilege dominating attention scores in $J$-stacking layers and sharply reduce the total space complexity to be $\bigO((2-\epsilon)L \log L)$, which helps receiving long sequence input.
\item We propose generative style decoder to acquire long sequence output with only one forward step needed, simultaneously avoiding cumulative error spreading during the inference phase.
\end{itemize}

\section{Preliminary}
\label{sec:method}

We first provide the LSTF problem definition.
Under the rolling forecasting setting with a fixed size window, we have the input $\mathcal{X}^t=\{\mb{x}^t_1,\ldots,\mb{x}^t_{L_x} ~|~ \mb{x}^t_{i} \in \mathbb{R}^{d_{x}}\}$ at time $t$, and the output is to predict corresponding sequence $\mathcal{Y}^t=\{\mb{y}^t_1,\ldots,\mb{y}^t_{L_y} ~|~ \mb{y}^t_{i} \in \mathbb{R}^{d_{y}} \}$. The LSTF problem encourages a longer output's length $L_y$ than previous works \cite{cho2014properties,sutskever2014sequence} and the feature dimension is not limited to univariate case ($d_{y} \geq 1$).

\textbf{Encoder-decoder architecture}
Many popular models are devised to ``encode'' the input representations $\mathcal{X}^t$ into a hidden state representations $\mathcal{H}^t$ and ``decode'' an output representations $\mathcal{Y}^t$ from $\mathcal{H}^t=\{\mb{h}^t_1,\ldots,\mb{h}^t_{L_h}\}$. The inference involves a step-by-step process named ``dynamic decoding'', where the decoder computes a new hidden state $\mb{h}^t_{k+1}$ from the previous state $\mb{h}^t_{k}$ and other necessary outputs from $k$-th step then predict the $(k+1)$-th sequence $\mb{y}^t_{k+1}$.

\textbf{Input Representation} A uniform input representation is given to enhance the global positional context and local temporal context of the time-series inputs. To avoid trivializing description, we put the details in Appendix B.

\begin{figure}
\centering
\includegraphics[width=\linewidth]{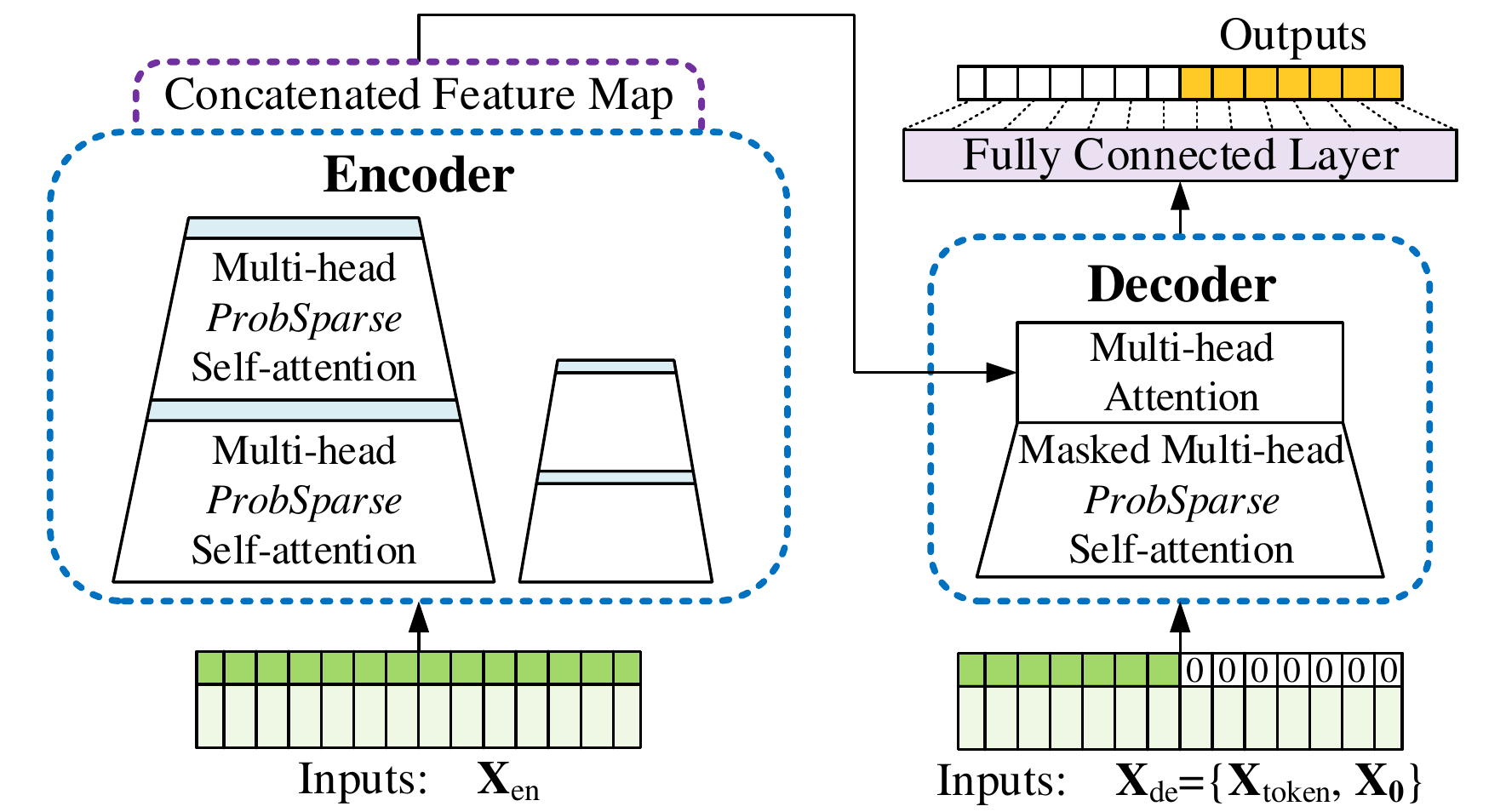}
\caption{{\mn} model overview. Left: The encoder receives massive long sequence inputs (green series). We replace canonical self-attention with the proposed \emph{ProbSparse} self-attention. The blue trapezoid is the self-attention distilling operation to extract dominating attention, reducing the network size sharply. The layer stacking replicas increase robustness. Right: The decoder receives long sequence inputs, pads the target elements into zero, measures the weighted attention composition of the feature map, and instantly predicts output elements (orange series) in a generative style.}
\label{fig:method.en.de.all}
\end{figure}

\section{Methodology}
Existing methods for time-series forecasting can be roughly grouped into two categories\footnote{Related work is in Appendix A due to space limitation.}. Classical time-series models serve as a reliable workhorse for time-series forecasting \cite{box2015time,ray1990time,seeger2017approximate,seeger2016bayesian}, and deep learning techniques mainly develop an encoder-decoder prediction paradigm by using RNN and their variants \cite{hochreiter1997long,li2017graph,yu2017long}.
Our proposed {\mn} holds the encoder-decoder architecture while targeting the LSTF problem. Please refer to Fig.(\ref{fig:method.en.de.all}) for an overview and the following sections for details.

\subsection{Efficient Self-attention Mechanism}
\label{sec:method.transformer.attention}

The canonical self-attention in \cite{vaswani2017attention} is defined based on the tuple inputs, i.e, query, key and value, which performs the scaled dot-product as $\mathcal{A} (\mb{Q}, \mb{K}, \mb{V}) = \textrm{Softmax} ({\mb{Q}\mb{K}^{\top}}/{\sqrt{d}})\mb{V}$,
where $\mb{Q} \in \mathbb{R}^{L_Q \times d}$, $\mb{K} \in \mathbb{R}^{L_K \times d}$, $\mb{V} \in \mathbb{R}^{L_V \times d}$ and $d$ is the input dimension.
To further discuss the self-attention mechanism, let $\mb{q}_{i}$, $\mb{k}_{i}$, $\mb{v}_{i}$ stand for the $i$-th row in $\mb{Q}$, $\mb{K}$, $\mb{V}$ respectively. Following the formulation in \cite{tsai2019transformer}, the $i$-th query's attention is defined as a kernel smoother in a probability form:
\begin{equation}\label{eq:method.prob.Attention}
    \mathcal{A} (\mb{q}_i, \mb{K}, \mb{V}) = \sum_{j} \frac{k(\mb{q}_i, \mb{k}_j)}{\sum_{l} k(\mb{q}_i, \mb{k}_l)} \mb{v}_{j} = \mathbb{E}_{p(\mb{k}_j|\mb{q}_i)} [\mb{v}_j] ~,
\end{equation}
where $p(\mb{k}_j|\mb{q}_i) = {k(\mb{q}_i, \mb{k}_j)}/{\sum_{l} k(\mb{q}_i, \mb{k}_l)}$ and $k(\mb{q}_i, \mb{k}_j)$ selects the asymmetric
exponential kernel $\exp ({\mb{q}_i \mb{k}_j^{\top}}/{\sqrt{d}})$.
The self-attention combines the values and acquires outputs based on computing the probability $p(\mb{k}_j|\mb{q}_i)$. It requires the quadratic times dot-product computation and $\bigO(L_Q L_K)$ memory usage, which is the major drawback when enhancing prediction capacity.

Some previous attempts have revealed that the distribution of self-attention probability has potential sparsity, and they have designed ``selective'' counting strategies on all $p(\mb{k}_j|\mb{q}_i)$ without significantly affecting the performance.
The Sparse Transformer \cite{child2019generating} incorporates both the row outputs and column inputs, in which the sparsity arises from the separated spatial correlation. The LogSparse Transformer \cite{li2019enhancing} notices the cyclical pattern in self-attention and forces each cell to attend to its previous one by an exponential step size.
The Longformer \cite{DBLP:journals/corr/abs-2004-05150} extends previous two works to more complicated sparse configuration.
However, they are limited to theoretical analysis from following heuristic methods and tackle each multi-head self-attention with the same strategy, which narrows their further improvement.

To motivate our approach, we first perform a qualitative assessment on the learned attention patterns of the canonical self-attention. The ``sparsity'' self-attention score forms a long tail distribution (see Appendix C for details), i.e., a few dot-product pairs contribute to the major attention, and others generate trivial attention. Then, the next question is how to distinguish them?

\begin{figure*}[t]
  \centering
  \includegraphics[width=0.85\linewidth]{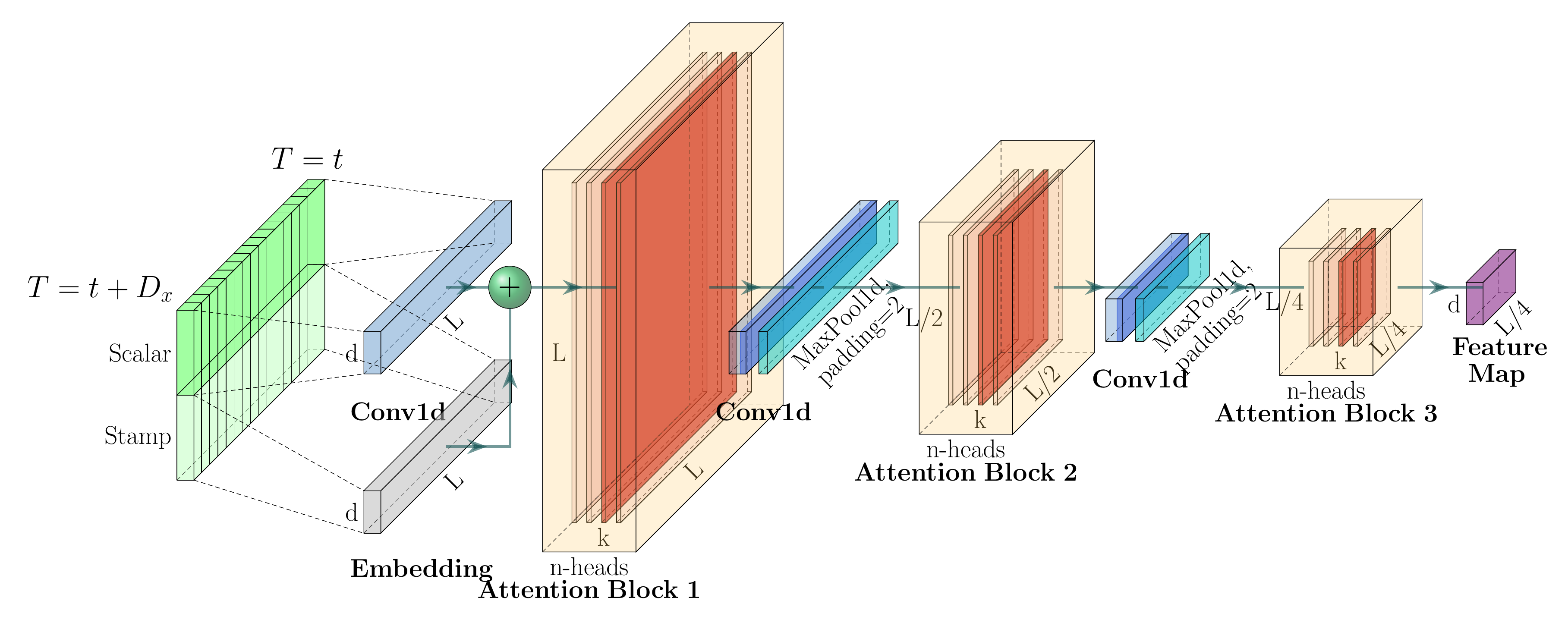}
  \caption{The single stack in \mn's encoder. (1) The horizontal stack stands for an individual one of the encoder replicas in Fig.(\ref{fig:method.en.de.all}). (2) The presented one is the main stack receiving the whole input sequence. Then the second stack takes half slices of the input, and the subsequent stacks repeat. (3) The red layers are dot-product matrixes, and they get cascade decrease by applying self-attention distilling on each layer. (4) Concatenate all stacks' feature maps as the encoder's output.}
  \label{fig:method.encoder}
\end{figure*}

\textbf{Query Sparsity Measurement}
From Eq.(\ref{eq:method.prob.Attention}), the $i$-th query's attention on all the keys are defined as a probability $p(\mb{k}_j|\mb{q}_i)$ and the output is its composition with values $\mb{v}$. The dominant dot-product pairs encourage the corresponding query's attention probability distribution away from the uniform distribution. If $p(\mb{k}_j|\mb{q}_i)$ is close to a uniform distribution $q(\mb{k}_j|\mb{q}_i)={1}/{L_{K}}$, the self-attention becomes a trivial sum of values $\mb{V}$ and is redundant to the residential input. Naturally, the ``likeness'' between distribution $p$ and $q$ can be used to distinguish the ``important'' queries. We measure the ``likeness'' through Kullback-Leibler divergence
$ KL (q||p) = \ln \sum_{l=1}^{L_K} e^{{\mb{q}_i\mb{k}_l^{\top}}/{\sqrt{d}}} - \frac{1}{L_K} \sum_{j=1}^{L_K} {\mb{q}_i\mb{k}_j^{\top}}/{\sqrt{d}} - \ln L_K $.
Dropping the constant, we define the $i$-th query's sparsity measurement as
\begin{equation}
\label{eq.method.measure}
    M (\mb{q}_i, \mb{K}) = \ln \sum_{j=1}^{L_K} e^{\frac{\mb{q}_i\mb{k}_j^{\top}}{\sqrt{d}}} - \frac{1}{L_K} \sum_{j=1}^{L_K} \frac{\mb{q}_i\mb{k}_j^{\top}}{\sqrt{d}} \qquad,
\end{equation}
where the first term is the Log-Sum-Exp (LSE) of $\mb{q}_i$ on all the keys, and the second term is the arithmetic mean on them. If the $i$-th query gains a larger $M(\mb{q}_i, \mb{K})$, its attention probability $p$ is more ``diverse'' and has a high chance to contain the dominate dot-product pairs in the header field of the long tail self-attention distribution.

\textbf{\emph{ProbSparse} Self-attention}
Based on the proposed measurement, we have the \emph{ProbSparse} self-attention by allowing each key to only attend to the $u$ dominant queries:
\begin{equation}
\label{eq:method.probsparse}
    \mathcal{A} (\mb{Q}, \mb{K}, \mb{V}) = \textrm{Softmax} (\frac{\overline{\mb{Q}}\mb{K}^{\top}}{\sqrt{d}})\mb{V} \qquad,
\end{equation}
where $\overline{\mb{Q}}$ is a sparse matrix of the same size of $\mb{q}$ and it only contains the Top-$u$ queries under the sparsity measurement $M(\mb{q},\mb{K})$. Controlled by a constant sampling factor $c$, we set $u=c \cdot \ln{L_Q}$, which makes the \emph{ProbSparse} self-attention only need to calculate $\bigO(\ln{L_Q})$ dot-product for each query-key lookup and the layer memory usage maintains $\bigO(L_K\ln{L_Q})$. Under the multi-head perspective, this attention generates different sparse query-key pairs for each head, which avoids severe information loss in return.

However, the traversing of all the queries for the measurement $M(\mb{q}_i,\mb{K})$ requires calculating each dot-product pairs, i.e., quadratically $\bigO(L_Q L_K)$, besides the LSE operation has the potential numerical stability issue. Motivated by this, we propose an empirical approximation for the efficient acquisition of the query sparsity measurement.

\begin{lemma}
\label{le.M.bound}
For each query $\mb{q}_i \in \mathbb{R}^{d}$ and $\mb{k}_j \in \mathbb{R}^{d}$ in the keys set $\mb{K}$, we have the bound as
$\ln L_K \leq M (\mb{q}_i, \mb{K}) \leq \max_{j}\{{\mb{q}_i \mb{k}_j^{\top}}/{\sqrt{d}}\} - \frac{1}{L_K} \sum_{j=1}^{L_K} \{{\mb{q}_i\mb{k}_j^{\top}}/{\sqrt{d}} \} + \ln L_K$.
When $\mb{q}_i \in \mb{K}$, it also holds.
\end{lemma}

From the Lemma \ref{le.M.bound} (proof is given in Appendix D.1), we propose the max-mean measurement as
\begin{equation}
\label{eq.method.measure2}
    \overline{M} (\mb{q}_i, \mb{K}) = \max_{j}\{\frac{\mb{q}_i \mb{k}_j^{\top}}{\sqrt{d}}\} - \frac{1}{L_K} \sum_{j=1}^{L_K} \frac{\mb{q}_i\mb{k}_j^{\top}}{\sqrt{d}} \quad.
\end{equation}

The range of Top-$u$ approximately holds in the boundary relaxation with Proposition 1 (refers in Appendix D.2). Under the long tail distribution, we only need to randomly sample $U=L_K \ln L_Q$ dot-product pairs to calculate the $\overline{M} (\mb{q}_i, \mb{K})$, i.e., filling other pairs with zero. Then, we select sparse Top-$u$ from them as $\overline{\mb{Q}}$. The max-operator in $\overline{M} (\mb{q}_i, \mb{K})$ is less sensitive to zero values and is numerical stable. In practice, the input length of queries and keys are typically equivalent in the self-attention computation, i.e $L_Q=L_K=L$ such that the total \emph{ProbSparse} self-attention time complexity and space complexity are $\bigO(L \ln L)$.

\subsection{Encoder: Allowing for Processing Longer Sequential Inputs under the Memory Usage Limitation}
\label{sec:method.transformer.encoder}
The encoder is designed to extract the robust long-range dependency of the long sequential inputs.
After the input representation, the $t$-th sequence input $\mathcal{X}^t$ has been shaped into a matrix $\mb{X}^t_{\textrm{en}} \in \mathbb{R}^{L_x \times d_{\textrm{model}}}$. We give a sketch of the encoder in Fig.(\ref{fig:method.encoder}) for clarity.

\textbf{Self-attention Distilling} As the natural consequence of the \emph{ProbSparse} self-attention mechanism, the encoder's feature map has redundant combinations of value $\mb{V}$. We use the distilling operation to privilege the superior ones with dominating features and make a focused self-attention feature map in the next layer. It trims the input's time dimension sharply, seeing the $n$-heads weights matrix (overlapping red squares) of Attention blocks in Fig.(\ref{fig:method.encoder}). Inspired by the dilated convolution \cite{yu2017dilated, gupta2017dilated}, our ``distilling'' procedure forwards from $j$-th layer into $(j+1)$-th layer as:
\begin{equation}
\label{eq:method.distilling}
\mb{X}^t_{j+1}=
    \mathrm{MaxPool}\left(~
    \mathrm{ELU}(~
    \mathrm{Conv1d}(
    [\mb{X}^t_{j}]_{\mathrm{AB}}
    )~)~\right) \quad, 
\end{equation}
where $[\cdot]_{\text{{AB}}}$ represents the attention block. It contains the Multi-head \emph{ProbSparse} self-attention and the essential operations, where $\textrm{Conv1d}(\cdot)$ performs an 1-D convolutional filters (kernel width=3) on time dimension with the $\textrm{ELU}(\cdot)$ activation function \cite{clevert2015fast}. We add a max-pooling layer with stride 2 and down-sample $\mb{X}^t$ into its half slice after stacking a layer, which reduces the whole memory usage to be $\bigO((2-\epsilon) L \log L)$, where $\epsilon$ is a small number. To enhance the robustness of the distilling operation, we build replicas of the main stack with halving inputs, and progressively decrease the number of self-attention distilling layers by dropping one layer at a time, like a pyramid in Fig.(\ref{fig:method.en.de.all}), such that their output dimension is aligned. Thus, we concatenate all the stacks' outputs and have the final hidden representation of encoder.

\subsection{Decoder: Generating Long Sequential Outputs Through One Forward Procedure}
\label{sec:method.decoder}
We use a standard decoder structure \cite{vaswani2017attention} in Fig.(\ref{fig:method.en.de.all}), and it is composed of a stack of two identical multi-head attention layers. However, the generative inference is employed to alleviate the speed plunge in long prediction. We feed the decoder with the following vectors as
\begin{equation}\label{eq.method.de_in}
    \mb{X}^t_{\textrm{de}} = \textrm{Concat} ( \mb{X}^t_{\textrm{token}}, \mb{X}^t_{\mb{0}}  ) \in \mathbb{R}^{(L_{\textrm{token}} + L_y) \times d_{\textrm{model}}} \quad,
\end{equation}
where $\mb{X}^t_{\textrm{token}} \in \mathbb{R}^{L_{\textrm{token}} \times d_{\textrm{model}}}$ is the start token, $\mb{X}^t_{\mb{0}} \in \mathbb{R}^{L_y \times d_{\textrm{model}}} $ is a placeholder for the target sequence (set scalar as 0).
Masked multi-head attention is applied in the \emph{ProbSparse} self-attention computing by setting masked dot-products to $-\infty$. It prevents each position from attending to coming positions, which avoids auto-regressive. A fully connected layer acquires the final output, and its outsize $d_y$ depends on whether we are performing a univariate forecasting or a multivariate one.

\textbf{Generative Inference} Start token is efficiently applied in NLP's ``dynamic decoding'' \cite{devlin2018bert}, and we extend it into a generative way. Instead of choosing specific flags as the token, we sample a $L_{\textrm{token}}$ long sequence in the input sequence, such as an earlier slice before the output sequence.
Take predicting 168 points as an example (7-day temperature prediction in the experiment section), we will take the known 5 days before the target sequence as ``start-token'', and feed the generative-style inference decoder with $\mb{X}_{\textrm{de}}=\{\mb{X}_{5d},\mb{X}_{\mb{0}}\}$. The $\mb{X}_{\mb{0}}$ contains target sequence's time stamp, i.e., the context at the target week.
Then our proposed decoder predicts outputs by one forward procedure rather than the time consuming ``dynamic decoding'' in the conventional encoder-decoder architecture. A detailed performance comparison is given in the computation efficiency section.

\textbf{Loss function}
We choose the MSE loss function on prediction w.r.t the target sequences, and the loss is propagated back from the decoder's outputs across the entire model.

\begin{table*}[t]
\centering
\fontsize{9pt}{9pt}\selectfont
\centering
\begin{tabular}{c|c|c|c|c|c|c|c|c|c}
\toprule[1.0pt]
\multicolumn{2}{c|}{Methods}         & {\mn} & {\mn$^{\dag}$} & {LogTrans} & {Reformer} & {LSTMa} & {DeepAR} & {ARIMA} & {Prophet} \\
\midrule[0.5pt]
\multicolumn{2}{c|}{Metric}          & MSE~~MAE               & MSE~~MAE                    & MSE~~MAE          & MSE~~MAE          & MSE~~MAE    & MSE~~MAE            & MSE~~MAE             & MSE~~MAE          \\
\midrule[1.0pt]
\multirow{5}{*}{\rotatebox{90}{ETTh$_1$}}	 & 	24	 & 	0.098~~0.247	 & 	\textbf{0.092}~~\textbf{0.246}	 & 	0.103~~0.259	 & 	0.222~~0.389	 & 	0.114~~0.272	 & 	0.107~~0.280	 & 	0.108~~0.284	 & 	0.115~~0.275\\
	 & 	48	 & 	\textbf{0.158}~~\textbf{0.319}	 & 	0.161~~0.322	 & 	0.167~~0.328	 & 	0.284~~0.445	 & 	0.193~~0.358	 & 	0.162~~0.327	 & 	0.175~~0.424	 & 	0.168~~0.330\\
	 & 	168	 & 	\textbf{0.183}~~\textbf{0.346}	 & 	0.187~~0.355	 & 	0.207~~0.375	 & 	1.522~~1.191	 & 	0.236~~0.392	 & 	0.239~~0.422	 & 	0.396~~0.504	 & 	1.224~~0.763\\
	 & 	336	 & 	0.222~~0.387	 & 	\textbf{0.215}~~\textbf{0.369}	 & 	0.230~~0.398	 & 	1.860~~1.124	 & 	0.590~~0.698	 & 	0.445~~0.552	 & 	0.468~~0.593	 & 	1.549~~1.820\\
	 & 	720	 & 	0.269~~0.435	 & 	\textbf{0.257}~~\textbf{0.421}	 & 	0.273~~0.463	 & 	2.112~~1.436	 & 	0.683~~0.768	 & 	0.658~~0.707	 & 	0.659~~0.766	 & 	2.735~~3.253\\
\midrule[0.5pt]
\multirow{5}{*}{\rotatebox{90}{ETTh$_2$}}	 & 	24	 & 	\textbf{0.093}~~\textbf{0.240}	 & 	0.099~~0.241	 & 	0.102~~0.255	 & 	0.263~~0.437	 & 	0.155~~0.307	 & 	0.098~~0.263	 & 	3.554~~0.445	 & 	0.199~~0.381\\
	 & 	48	 & 	\textbf{0.155}~~\textbf{0.314}	 & 	0.159~~0.317	 & 	0.169~~0.348	 & 	0.458~~0.545	 & 	0.190~~0.348	 & 	0.163~~0.341	 & 	3.190~~0.474	 & 	0.304~~0.462\\
	 & 	168	 & 	\textbf{0.232}~~\textbf{0.389}	 & 	0.235~~0.390	 & 	0.246~~0.422	 & 	1.029~~0.879	 & 	0.385~~0.514	 & 	0.255~~0.414	 & 	2.800~~0.595	 & 	2.145~~1.068\\
	 & 	336	 & 	0.263~~\textbf{0.417}	 & 	\textbf{0.258}~~0.423	 & 	0.267~~0.437	 & 	1.668~~1.228	 & 	0.558~~0.606	 & 	0.604~~0.607	 & 	2.753~~0.738	 & 	2.096~~2.543\\
	 & 	720	 & 	\textbf{0.277}~~\textbf{ 0.431}	 & 	0.285~~0.442	 & 	0.303~~0.493	 & 	2.030~~1.721	 & 	0.640~~0.681	 & 	0.429~~0.580	 & 	2.878~~1.044	 & 	3.355~~4.664\\
\midrule[0.5pt]
\multirow{5}{*}{\rotatebox{90}{ETTm$_1$}}	 & 	24	 & 	\textbf{0.030}~~\textbf{0.137}	 & 	0.034~~0.160	 & 	0.065~~0.202	 & 	0.095~~0.228	 & 	0.121~~0.233	 & 	0.091~~0.243	 & 	0.090~~0.206	 & 	0.120~~0.290\\
	 & 	48	 & 	0.069~~0.203	 & 	\textbf{0.066}~~\textbf{0.194}	 & 	0.078~~0.220	 & 	0.249~~0.390	 & 	0.305~~0.411	 & 	0.219~~0.362	 & 	0.179~~0.306	 & 	0.133~~0.305\\
	 & 	96	 & 	0.194~~\textbf{0.372}	 & 	\textbf{0.187}~~0.384	 & 	0.199~~0.386	 & 	0.920~~0.767	 & 	0.287~~0.420	 & 	0.364~~0.496	 & 	0.272~~0.399	 & 	0.194~~0.396\\
	 & 	288	 & 	\textbf{0.401}~~0.554	 & 	0.409~~\textbf{0.548}	 & 	0.411~~0.572	 & 	1.108~~1.245	 & 	0.524~~0.584	 & 	0.948~~0.795	 & 	0.462~~0.558	 & 	0.452~~0.574\\
	 & 	672	 & 	\textbf{0.512}~~\textbf{0.644}	 & 	0.519~~0.665	 & 	0.598~~0.702	 & 	1.793~~1.528	 & 	1.064~~0.873	 & 	2.437~~1.352	 & 	0.639~~0.697	 & 	2.747~~1.174\\
\midrule[0.5pt]
\multirow{5}{*}{\rotatebox{90}{Weather}}	 & 	24	 & 	\textbf{0.117}~~\textbf{0.251}	 & 	0.119~~0.256	 & 	0.136~~0.279	 & 	0.231~~0.401	 & 	0.131~~0.254	 & 	0.128~~0.274	 & 	0.219~~0.355	 & 	0.302~~0.433\\
	 & 	48	 & 	\textbf{0.178}~~0.318	 & 	0.185~~\textbf{0.316}	 & 	0.206~~0.356	 & 	0.328~~0.423	 & 	0.190~~0.334	 & 	0.203~~0.353	 & 	0.273~~0.409	 & 	0.445~~0.536\\
	 & 	168	 & 	\textbf{0.266}~~\textbf{0.398}	 & 	0.269~~0.404	 & 	0.309~~0.439	 & 	0.654~~0.634	 & 	0.341~~0.448	 & 	0.293~~0.451	 & 	0.503~~0.599	 & 	2.441~~1.142\\
	 & 	336	 & 	\textbf{0.297}~~\textbf{0.416}	 & 	0.310~~0.422	 & 	0.359~~0.484	 & 	1.792~~1.093	 & 	0.456~~0.554	 & 	0.585~~0.644	 & 	0.728~~0.730	 & 	1.987~~2.468\\
	 & 	720	 & 	\textbf{0.359}~~\textbf{0.466}	 & 	0.361~~0.471	 & 	0.388~~0.499	 & 	2.087~~1.534	 & 	0.866~~0.809	 & 	0.499~~0.596	 & 	1.062~~0.943	 & 	3.859~~1.144\\
\midrule[0.5pt]
\multirow{5}{*}{\rotatebox{90}{ECL}}	 & 	48	 & 	0.239~~0.359	 & 	0.238~~0.368	 & 	0.280~~0.429	 & 	0.971~~0.884	 & 	0.493~~0.539	 & 	\textbf{0.204}~~\textbf{0.357}	 & 	0.879~~0.764	 & 	0.524~~0.595\\
	 & 	168	 & 	0.447~~0.503	 & 	0.442~~0.514	 & 	0.454~~0.529	 & 	1.671~~1.587	 & 	0.723~~0.655	 & 	\textbf{0.315}~~\textbf{0.436}	 & 	1.032~~0.833	 & 	2.725~~1.273\\
	 & 	336	 & 	0.489~~0.528	 & 	0.501~~0.552	 & 	0.514~~0.563	 & 	3.528~~2.196	 & 	1.212~~0.898	 & 	\textbf{0.414}~~\textbf{0.519}	 & 	1.136~~0.876	 & 	2.246~~3.077\\
	 & 	720	 & 	\textbf{0.540}~~\textbf{0.571}	 & 	0.543~~0.578	 & 	0.558~~0.609	 & 	4.891~~4.047	 & 	1.511~~0.966	 & 	0.563~~0.595	 & 	1.251~~0.933	 & 	4.243~~1.415\\
	 & 	960	 & 	\textbf{0.582}~~\textbf{0.608}	 & 	0.594~~0.638	 & 	0.624~~0.645	 & 	7.019~~5.105	 & 	1.545~~1.006	 & 	0.657~~0.683	 & 	1.370~~0.982	 & 	6.901~~4.264\\

\midrule[1.0pt]
\multicolumn{2}{c|}{Count}           & {32}                 & {12}                          & {0}        & {0}        & {0}     & {6}           & {0}     & {0}      \\
\bottomrule[1.0pt]

\end{tabular}
\caption{Univariate long sequence time-series forecasting results on four datasets (five cases).}
\label{tab:exp.mainResults}
\end{table*}

\section{Experiment}
\label{sec:exp}

\subsection{Datasets}
We extensively perform experiments on four datasets, including 2 collected real-world datasets for LSTF and 2 public benchmark datasets.

\textbf{ETT} (Electricity Transformer Temperature)\footnote{We collected the ETT dataset and published it at \url{https://github.com/zhouhaoyi/ETDataset}.}:
The ETT is a crucial indicator in the electric power long-term deployment.
We collected 2-year data from two separated counties in China. To explore the granularity on the LSTF problem, we create separate datasets as \{ETTh$_1$, ETTh$_2$\} for 1-hour-level and ETTm$_1$ for 15-minute-level.
Each data point consists of the target value "oil temperature" and 6 power load features.
The train/val/test is 12/4/4 months.

\textbf{ECL} (Electricity Consuming Load)\footnote{ECL dataset was acquired at \url{https://archive.ics.uci.edu/ml/datasets/ElectricityLoadDiagrams20112014}.}: It collects the electricity consumption (Kwh) of 321 clients. Due to the missing data \cite{li2019enhancing}, we convert the dataset into hourly consumption of 2 years and set `MT\_320' as the target value. The train/val/test is 15/3/4 months.

\textbf{Weather} \footnote{Weather dataset was acquired at \url{https://www.ncei.noaa.gov/data/local-climatological-data/}.}: This dataset contains local climatological data for nearly 1,600 U.S. locations, 4 years from 2010 to 2013, where data points are collected every 1 hour. Each data point consists of the target value ``wet bulb" and 11 climate features. The train/val/test is 28/10/10 months.


\begin{table*}[t]
\centering
\fontsize{9pt}{9pt}\selectfont
\begin{tabular}{c|c|cc|cc|cc|cc|cc|cc}
\toprule[1.0pt]
\multicolumn{2}{c}{Methods}     & \multicolumn{2}{|c}{\mn} & \multicolumn{2}{|c}{\mn$^{\dag}$} & \multicolumn{2}{|c}{LogTrans} & \multicolumn{2}{|c}{Reformer} & \multicolumn{2}{|c}{LSTMa} & \multicolumn{2}{|c}{LSTnet}      \\
\midrule[0.5pt]
\multicolumn{2}{c|}{Metric}      & MSE                & MAE               & MSE                          & MAE              & MSE               & MAE      & MSE           & MAE          & MSE         & MAE         & MSE            & MAE            \\
\midrule[1.0pt]
\multirow{5}{*}{\rotatebox{90}{ETTh$_1$}} & 24  & \textbf{0.577}          & \textbf{0.549}          & 0.620                   & 0.577                   & 0.686                   & 0.604                   & 0.991                   & 0.754                   & 0.650                   & 0.624                   & 1.293                   & 0.901                   \\
                          & 48  & \textbf{0.685}          & \textbf{0.625}          & 0.692                   & 0.671                   & 0.766                   & 0.757                   & 1.313                   & 0.906                   & 0.702                   & 0.675                   & 1.456                   & 0.960                   \\
                          & 168 & \textbf{0.931}          & \textbf{0.752}          & 0.947                   & 0.797                   & 1.002                   & 0.846                   & 1.824                   & 1.138                   & 1.212                   & 0.867                   & 1.997                   & 1.214                   \\
                          & 336 & 1.128                   & 0.873                   & \textbf{1.094}          & \textbf{0.813}          & 1.362                   & 0.952                   & 2.117                   & 1.280                   & 1.424                   & 0.994                   & 2.655                   & 1.369                   \\
                          & 720 & \textbf{1.215}          & \textbf{0.896}          & 1.241                   & 0.917                   & 1.397                   & 1.291                   & 2.415                   & 1.520                   & 1.960                   & 1.322                   & 2.143                   & 1.380                   \\
\midrule[0.5pt]
\multirow{5}{*}{\rotatebox{90}{ETTh$_2$}} & 24  & \textbf{0.720}          & \textbf{0.665}          & 0.753                   & 0.727                   & 0.828                   & 0.750                   & 1.531                   & 1.613                   & 1.143                   & 0.813                   & 2.742                   & 1.457                   \\
                          & 48  & \textbf{1.457}          & \textbf{1.001}          & 1.461                   & 1.077                   & 1.806                   & 1.034                   & 1.871                   & 1.735                   & 1.671                   & 1.221                   & 3.567                   & 1.687                   \\
                          & 168 & 3.489                   & \textbf{1.515}          & 3.485                   & 1.612                   & 4.070                   & 1.681                   & 4.660                   & 1.846                   & 4.117                   & 1.674                   & \textbf{3.242}          & 2.513                   \\
                          & 336 & 2.723                   & 1.340                   & 2.626                   & \textbf{1.285}          & 3.875                   & 1.763                   & 4.028                   & 1.688                   & 3.434                   & 1.549                   & \textbf{2.544}          & 2.591                   \\
                          & 720 & \textbf{3.467}          & \textbf{1.473}          & 3.548                   & 1.495                   & 3.913                   & 1.552                   & 5.381                   & 2.015                   & 3.963                   & 1.788                   & 4.625                   & 3.709                   \\
\midrule[0.5pt]
\multirow{5}{*}{\rotatebox{90}{ETTm$_1$}} & 24  & 0.323                   & \textbf{0.369}          & \textbf{0.306}          & 0.371                   & 0.419                   & 0.412                   & 0.724                   & 0.607                   & 0.621                   & 0.629                   & 1.968                   & 1.170                   \\
                          & 48  & 0.494                   & 0.503                   & \textbf{0.465}          & \textbf{0.470}          & 0.507                   & 0.583                   & 1.098                   & 0.777                   & 1.392                   & 0.939                   & 1.999                   & 1.215                   \\
                          & 96  & \textbf{0.678}          & 0.614                   & 0.681                   & \textbf{0.612}          & 0.768                   & 0.792                   & 1.433                   & 0.945                   & 1.339                   & 0.913                   & 2.762                   & 1.542                   \\
                          & 288 & \textbf{1.056}                   & \textbf{0.786}          & 1.162                   & 0.879                   & 1.462                   & 1.320                   & 1.820                   & 1.094                   & 1.740                   & 1.124                   & 1.257          & 2.076                   \\
                          & 672 & \textbf{1.192}          & \textbf{0.926}          & 1.231                   & 1.103                   & 1.669                   & 1.461                   & 2.187                   & 1.232                   & 2.736                   & 1.555                   & 1.917                   & 2.941                   \\
\midrule[0.5pt]
\multirow{5}{*}{\rotatebox{90}{Weather}}  & 24  & \textbf{0.335}          & \textbf{0.381}          & 0.349                   & 0.397                   & 0.435                   & 0.477                   & 0.655                   & 0.583                   & 0.546                   & 0.570                   & 0.615                   & 0.545                   \\
                          & 48  & 0.395                   & 0.459                   & \textbf{0.386}          & \textbf{0.433}          & 0.426                   & 0.495                   & 0.729                   & 0.666                   & 0.829                   & 0.677                   & 0.660                   & 0.589                   \\
                          & 168 & \textbf{0.608}          & \textbf{0.567}          & 0.613                   & 0.582                   & 0.727                   & 0.671                   & 1.318                   & 0.855                   & 1.038                   & 0.835                   & 0.748                   & 0.647                   \\
                          & 336 & \textbf{0.702}          & \textbf{0.620}          & 0.707                   & 0.634                   & 0.754                   & 0.670                   & 1.930                   & 1.167                   & 1.657                   & 1.059                   & 0.782                   & 0.683                   \\
                          & 720 & \textbf{0.831}         & \textbf{0.731}          & 0.834                   & 0.741                   & 0.885                   & 0.773                   & 2.726                   & 1.575                   & 1.536                   & 1.109                   & 0.851                   & 0.757                   \\
\midrule[0.5pt]
\multirow{5}{*}{\rotatebox{90}{ECL}}      & 48  & 0.344                   & \textbf{0.393}          & \textbf{0.334}          & 0.399                   & 0.355                   & 0.418                   & 1.404                   & 0.999                   & 0.486                   & 0.572                   & 0.369                   & 0.445                   \\
                          & 168 & 0.368                   & 0.424          & \textbf{0.353}          & \textbf{0.420}                   & 0.368                   & 0.432                   & 1.515                   & 1.069                   & 0.574                   & 0.602                   & 0.394                   & 0.476                   \\
                          & 336 & 0.381                   & \textbf{0.431}          & 0.381                   & 0.439                   & \textbf{0.373}          & 0.439                   & 1.601                   & 1.104                   & 0.886                   & 0.795                   & 0.419                   & 0.477                   \\
                          & 720 & 0.406                   & 0.443                   & \textbf{0.391}          & \textbf{0.438}          & 0.409                   & 0.454                   & 2.009                   & 1.170                   & 1.676                   & 1.095                   & 0.556                   & 0.565                   \\
                          & 960 & \textbf{0.460}          & \textbf{0.548}          & 0.492                   & 0.550                   & 0.477                   & 0.589                   & 2.141                   & 1.387                   & 1.591                   & 1.128                   & 0.605                   & 0.599                   \\
\midrule[1.0pt]
\multicolumn{2}{c}{Count}       & \multicolumn{2}{|c}{33}                 & \multicolumn{2}{|c}{14}                          & \multicolumn{2}{|c}{1}        & \multicolumn{2}{|c}{0}        & \multicolumn{2}{|c}{0}     & \multicolumn{2}{|c}{2}    \\
\bottomrule[1.0pt]

\end{tabular}
\caption{Multivariate long sequence time-series forecasting results on four datasets (five cases).}
\label{tab:exp.multivarResults}
\end{table*}

\subsection{Experimental Details}
We briefly summarize basics, and more information on network components and setups are given in Appendix E.

\textbf{Baselines:}
We have selected five time-series forecasting methods as comparison, including ARIMA \cite{ariyo2014stock}, Prophet \cite{taylor2018forecasting}, LSTMa \cite{bahdanau2014neural}, LSTnet \cite{lai2018modeling} and DeepAR \cite{flunkert2017deepar}. To better explore the \emph{ProbSparse} self-attention's performance in our proposed {\mn}, we incorporate the canonical self-attention variant (\mn$^{\dag}$), the efficient variant Reformer \cite{kitaev2019reformer} and the most related work LogSparse self-attention \cite{li2019enhancing} in the experiments.
The details of network components are given in Appendix E.1. 

\textbf{Hyper-parameter tuning:}
We conduct grid search over the hyper-parameters, and detailed ranges are given in Appendix E.3. {\mn} contains a 3-layer stack and a 1-layer stack (1/4 input) in the encoder, and a 2-layer decoder. Our proposed methods are optimized with Adam optimizer, and its learning rate starts from $1e^{-4}$, decaying two times smaller every epoch. The total number of epochs is 8 with proper early stopping. We set the comparison methods as recommended, and the batch size is 32.
\textbf{Setup:}
The input of each dataset is zero-mean normalized.
Under the LSTF settings, we prolong the prediction windows size $L_y$ progressively, i.e., \{1d, 2d, 7d, 14d, 30d, 40d\} in \{ETTh, ECL, Weather\}, \{6h, 12h, 24h, 72h, 168h\} in ETTm.
\textbf{Metrics:} We use two evaluation metrics, including $\textrm{MSE}=\frac{1}{n}\sum_{i=1}^{n}(\mb{y}-\hat{\mb{y}})^2$ and $\textrm{MAE}=\frac{1}{n}\sum_{i=1}^{n}|\mb{y}-\hat{\mb{y}}|$ on each prediction window (averaging for multivariate prediction), and roll the whole set with $\textrm{stride}=1$.
\textbf{Platform:} All the models were trained/tested on a single Nvidia V100 32GB GPU. The source code is available at \url{https://github.com/zhouhaoyi/Informer2020}.

\begin{figure*}[t]
\centering
    \subfigure[Input length.]{
    \includegraphics[height=0.23\linewidth]{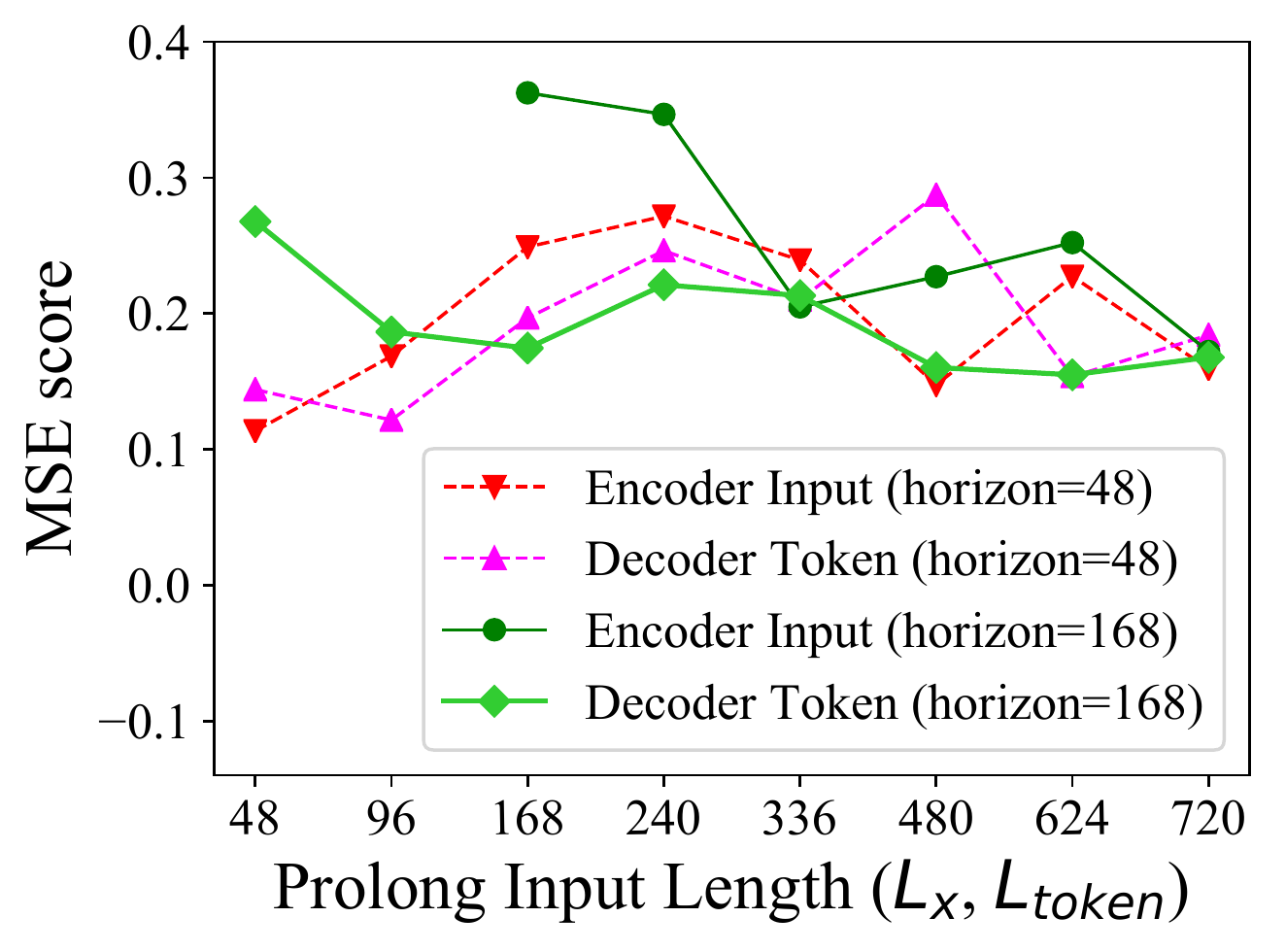}
    \label{fig:sensi_length}
    }
    \hfill
    \subfigure[Sampling Factor.]{
    \includegraphics[height=0.23\linewidth]{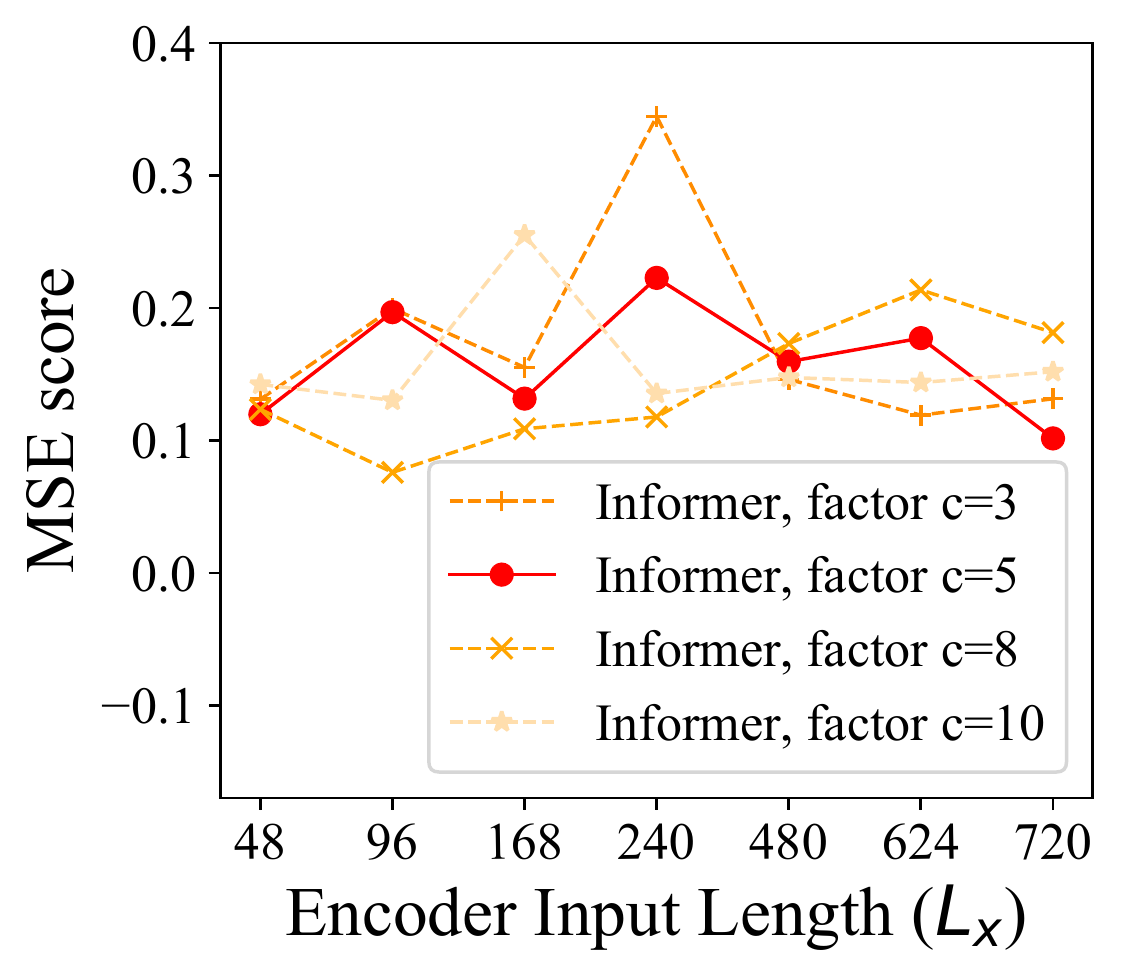}
    \label{fig:sensi_factor}
    }
    \hfill
    \subfigure[Stacking Combination.]{
    \includegraphics[height=0.23\linewidth]{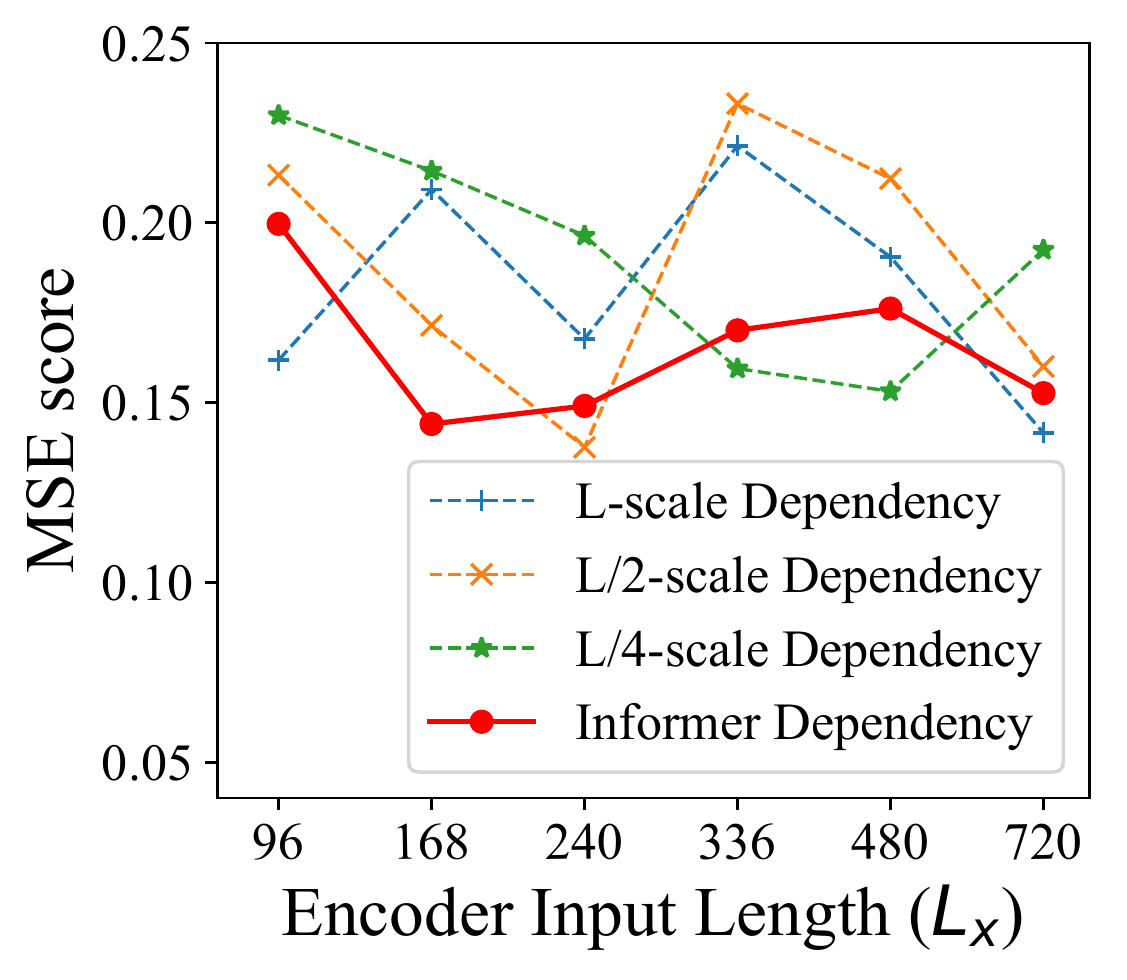}
    \label{fig:sensi_stacks}
    }
\caption{The parameter sensitivity of three components in \mn. \qquad\qquad~~~}
\label{fig:sensi}
\end{figure*}

\begin{table*}[t]
\begin{minipage}{0.58\linewidth}
\fontsize{9pt}{9pt}\selectfont
\begin{threeparttable}[b]
\begin{tabular}{cc|ccc|ccc}
\toprule[1.0pt]
\multicolumn{2}{l|}{Prediction length} & \multicolumn{3}{c|}{336} & \multicolumn{3}{c}{720}  \\
\midrule[0.5pt]
\multicolumn{2}{l|}{Encoder's input}   & 336    & 720    & 1440   & 720    & 1440   & 2880   \\
\midrule[1.0pt]
\multirow{2}{*}{\mn}       & MSE       & 0.249 & 0.225 & 0.216 & 0.271 & 0.261 & 0.257 \\
                           & MAE       & 0.393 & 0.384 & 0.376 & 0.435 & 0.431 & 0.422 \\
\midrule[0.5pt]
    \multirow{2}{*}{\mn$^{\dag}$}       & MSE       & 0.241 & 0.214 & - & 0.259 & - & - \\
                           & MAE       & 0.383 & 0.371 & - & 0.423 & - & - \\
\midrule[0.5pt]
\multirow{2}{*}{LogTrans}       & MSE       & 0.263 & 0.231 & - & 0.273 & - & -      \\
                           & MAE       & 0.418 & 0.398 & - & 0.463 & - & -      \\
\midrule[0.5pt]
\multirow{2}{*}{Reformer}       & MSE       & 1.875 & 1.865 & 1.861 & 2.243 & 2.174 & 2.113      \\
                           & MAE       & 1.144 & 1.129 & 1.125 & 1.536 & 1.497 & 1.434      \\
\bottomrule[1.0pt]
\end{tabular}
\begin{tablenotes}
\item[1] \mn$^{\dag}$ uses the canonical self-attention mechanism.
\item[2] The `-' indicates failure for the out-of-memory.
\end{tablenotes}
\caption{Ablation study of the \emph{ProbSparse} self-attention mechanism.}
\label{tab:exp.ablation.sparse}
\end{threeparttable}
\end{minipage}
\begin{minipage}{0.41\linewidth}
\centering
\fontsize{9pt}{9pt}\selectfont
\begin{threeparttable}
\begin{tabular}{l|c|c|c}
\toprule[1.0pt]
\multirow{3}{*}{Methods}  & \multicolumn{2}{c|}{Training} & Testing     \\
\cmidrule{2-4}
         & Time          & Memory  & Steps             \\
\midrule[1.0pt]
\mn   & $\bigO (L \log L)$   & $\bigO (L \log L)$ & 1                       \\
\midrule[0.5pt]
Transformer   & $\bigO (L^2)$   & $\bigO (L^2)$ & $L$                       \\
\midrule[0.5pt]
LogTrans & $\bigO (L \log L)$   & $\bigO (L^2)$ & 1$^{\star}$                      \\
\midrule[0.5pt]
Reformer & $\bigO (L \log L)$   & $\bigO (L \log L)$ & $L$          \\
\midrule[0.5pt]
LSTM     & $\bigO (L)$   & $\bigO (L)$ & $L$                     \\
\bottomrule[1.0pt]
\end{tabular}
\begin{tablenotes}
\item[1] The LSTnet is hard to present in a closed form.
\item[2] The ${\star}$ denotes applying our proposed decoder. 
\end{tablenotes}
\end{threeparttable}
\captionsetup{type=table} 
\caption{$L$-related computation statics of each layer.}
\label{tab:exp.computation.summary}
\end{minipage}
\end{table*}

\subsection{Results and Analysis}
\label{sec:exp.overallresult}
Table \ref{tab:exp.mainResults} and Table \ref{tab:exp.multivarResults} summarize the univariate/multivariate evaluation results of all the methods on 4 datasets. We gradually prolong the prediction horizon as a higher requirement of prediction capacity, where the LSTF problem setting is precisely controlled to be tractable on one single GPU for each method.
The best results are highlighted in boldface.

\textbf{Univariate Time-series Forecasting}
Under this setting, each method attains predictions as a single variable over time series. From Table \ref{tab:exp.mainResults}, we can observe that:
\textbf{(1)} The proposed model {\mn} significantly improves the inference performance (wining-counts in the last column) across all datasets, and their predict error rises smoothly and slowly within the growing prediction horizon, which demonstrates the success of {\mn} in enhancing the prediction capacity in the LSTF problem.
\textbf{(2)} The {\mn} beats its canonical degradation \mn$^{\dag}$ mostly in wining-counts, i.e., 32$>$12, which supports the query sparsity assumption in providing a comparable attention feature map. Our proposed method also out-performs the most related work LogTrans and Reformer. We note that the Reformer keeps dynamic decoding and performs poorly in LSTF, while other methods benefit from the generative style decoder as nonautoregressive predictors.
\textbf{(3)} The {\mn} model shows significantly better results than recurrent neural networks LSTMa. Our method has a MSE decrease of 26.8\% (at 168), 52.4\% (at 336) and 60.1\% (at 720).
This reveals a shorter network path in the self-attention mechanism acquires better prediction capacity than the RNN-based models.
\textbf{(4)} The proposed method outperforms DeepAR, ARIMA and Prophet on MSE by decreasing 49.3\% (at 168), 61.1\% (at 336), and 65.1\% (at 720) in average.
On the ECL dataset, DeepAR performs better on shorter horizons ($\leq 336$), and our method surpasses on longer horizons. We attribute this to a specific example, in which the effectiveness of prediction capacity is reflected with the problem scalability.

\textbf{Multivariate Time-series Forecasting}
Within this setting, some univariate methods are inappropriate, and LSTnet is the state-of-art baseline. On the contrary, our proposed {\mn} is easy to change from univariate prediction to multivariate one by adjusting the final FCN layer.
From Table \ref{tab:exp.multivarResults}, we observe that: \textbf{(1)} The proposed model {\mn} greatly outperforms other methods and the findings 1 \& 2 in the univariate settings still hold for the multivariate time-series.
\textbf{(2)} The {\mn} model shows better results than RNN-based LSTMa and CNN-based LSTnet, and the MSE decreases 26.6\% (at 168), 28.2\% (at 336), 34.3\% (at 720) in average. Compared with the univariate results, the overwhelming performance is reduced, and such phenomena can be caused by the anisotropy of feature dimensions' prediction capacity. It is beyond the scope of this paper, and we will explore it in the future work.


\textbf{LSTF with Granularity Consideration}
We perform an additional comparison to explore the performance with various granularities. The sequences \{96, 288, 672\} of ETTm$_1$ (minutes-level) are aligned with \{24, 48, 168\} of ETTh$_1$ (hour-level). The {\mn} outperforms other baselines even if the sequences are at different granularity levels.

\subsection{Parameter Sensitivity}
We perform the sensitivity analysis of the proposed {\mn} model on ETTh1 under the univariate setting.
\textbf{Input Length:} In Fig.(\ref{fig:sensi}a), when predicting short sequences (like 48), initially increasing input length of encoder/decoder degrades performance, but further increasing causes the MSE to drop because it brings repeat short-term patterns. However, the MSE gets lower with longer inputs in predicting long sequences (like 168). Because the longer encoder input may contain more dependencies, and the longer decoder token has rich local information.
\textbf{Sampling Factor:} The sampling factor controls the information bandwidth of \emph{ProbSparse} self-attention in Eq.(\ref{eq:method.probsparse}). We start from the small factor (=3) to large ones, and the general performance increases a little and stabilizes at last in Fig.(\ref{fig:sensi}b). It verifies our query sparsity assumption that there are redundant dot-product pairs in the self-attention mechanism. We set the sample factor $c=5$ (the red line) in practice.
\textbf{The Combination of Layer Stacking:} The replica of Layers is complementary for the self-attention distilling, and we investigate each stack \{L, L/2, L/4\}'s behavior in Fig.(\ref{fig:sensi}c). The longer stack is more sensitive to the inputs, partly due to receiving more long-term information. Our method's selection (the red line), i.e., joining L and L/4, is the most robust strategy.

\subsection{Ablation Study: How well {\mn} works?}
We also conducted additional experiments on ETTh$_1$ with ablation consideration.

\begin{table*}[t]
\centering
\fontsize{9pt}{9pt}\selectfont
\begin{threeparttable}
\begin{tabular}{cc|ccccc|ccccc}
\toprule[1.0pt]
\multicolumn{2}{l|}{Prediction length}      & \multicolumn{5}{c|}{336}           & \multicolumn{5}{c}{480} \\
\midrule[0.5pt]
\multicolumn{2}{l|}{Encoder's input} & 336 & 480 & 720 & 960 & 1200 & 336 & 480 & 720 & 960 & 1200\\
\midrule[1.0pt]
\multirow{2}{*}{\mn$^{\dag}$} & MSE & 0.249 & 0.208 & 0.225 & 0.199 & 0.186 & 0.197 & 0.243 & 0.213 & 0.192 & 0.174\\
                              & MAE & 0.393 & 0.385 & 0.384 & 0.371 & 0.365 & 0.388 & 0.392 & 0.383 & 0.377 & 0.362 \\
\midrule[0.5pt]
\multirow{2}{*}{\mn$^{\ddag}$} & MSE & 0.229 & 0.215 & 0.204 & - & - & 0.224 & 0.208 & 0.197 & - & -  \\
                               & MAE & 0.391 & 0.387 & 0.377 & - & - & 0.381 & 0.376 & 0.370 & - & -\\
\bottomrule[1.0pt]
\end{tabular}
\begin{tablenotes}
\item[1] \mn$^{\ddag}$ removes the self-attention distilling from \mn$^{\dag}$.
\item[2] The `-' indicates failure for the out-of-memory.
\end{tablenotes}
\end{threeparttable}
\caption{Ablation study of the self-attention distilling.}
\label{tab:exp.ablation.distilling}
\end{table*}

\begin{table*}[t]
\centering
\fontsize{9pt}{9pt}\selectfont
\begin{threeparttable}
\begin{tabular}{cc|ccccc|ccccc}
\toprule[1.0pt]
\multicolumn{2}{l|}{Prediction length} & \multicolumn{5}{c|}{336} & \multicolumn{5}{c}{480}  \\
\midrule[0.5pt]
\multicolumn{2}{l|}{Prediction offset} & +0 & +12 & +24 & +48 & +72 & +0 & +48 & +96 & +144 & +168 \\ \midrule[1.0pt]
\multirow{2}{*}{\mn$^{\ddag}$} & MSE & 0.207 & 0.209 & 0.211 & 0.211 & 0.216 & 0.198 & 0.203 & 0.203 & 0.208 & 0.208 \\
                     & MAE & 0.385 & 0.387 & 0.391 & 0.393 & 0.397 & 0.390 & 0.392 & 0.393 & 0.401 & 0.403\\
\midrule[0.5pt]
\multirow{2}{*}{\mn$^{\S}$} & MSE & 0.201 & - & - & - & - & 0.392 & - & - & - & - \\
                     & MAE & 0.393 & - & - & - & - & 0.484 & - & - & - & - \\
\bottomrule[1.0pt]
\end{tabular}
\begin{tablenotes}
\item[1] \mn$^{\S}$ replaces our decoder with dynamic decoding one in \mn$^{\ddag}$.
\item[2] The `-' indicates failure for the unacceptable metric results.
\end{tablenotes}
\end{threeparttable}
\caption{Ablation study of the generative style decoder.}
\label{tab:exp.ablation.decoder}
\vspace{-1 ex}
\end{table*}

\textbf{The performance of \emph{ProbSparse} self-attention mechanism}
In the overall results Table \ref{tab:exp.mainResults} \& \ref{tab:exp.multivarResults}, we limited the problem setting to make the memory usage feasible for the canonical self-attention.
In this study, we compare our methods with LogTrans and Reformer, and thoroughly explore their extreme performance.
To isolate the memory efficient problem, we first reduce settings as \{batch size=8, heads=8, dim=64\}, and maintain other setups in the univariate case. 
In Table \ref{tab:exp.ablation.sparse}, the \emph{ProbSparse} self-attention shows better performance than the counterparts. The LogTrans gets OOM in extreme cases because its public implementation is the mask of the full-attention, which still has $\bigO(L^2)$ memory usage. Our proposed \emph{ProbSparse} self-attention avoids this from the simplicity brought by the query sparsity assumption in Eq.(\ref{eq.method.measure2}), referring to the pseudo-code in Appendix E.2, and reaches smaller memory usage.

\textbf{The performance of self-attention distilling}
In this study, we use \mn$^{\dag}$ as the benchmark to eliminate additional effects of \emph{ProbSparse} self-attention.
The other experimental setup is aligned with the settings of univariate Time-series.
From Table \ref{tab:exp.ablation.distilling}, \mn$^{\dag}$ has fulfilled all the experiments and achieves better performance after taking advantage of long sequence inputs. The comparison method \mn$^{\ddag}$ removes the distilling operation and reaches OOM with longer inputs ($> 720$).
Regarding the benefits of long sequence inputs in the LSTF problem, we conclude that the self-attention distilling is worth adopting, especially when a longer prediction is required.

\begin{figure}[t]
  \centering
  \includegraphics[width=\linewidth]{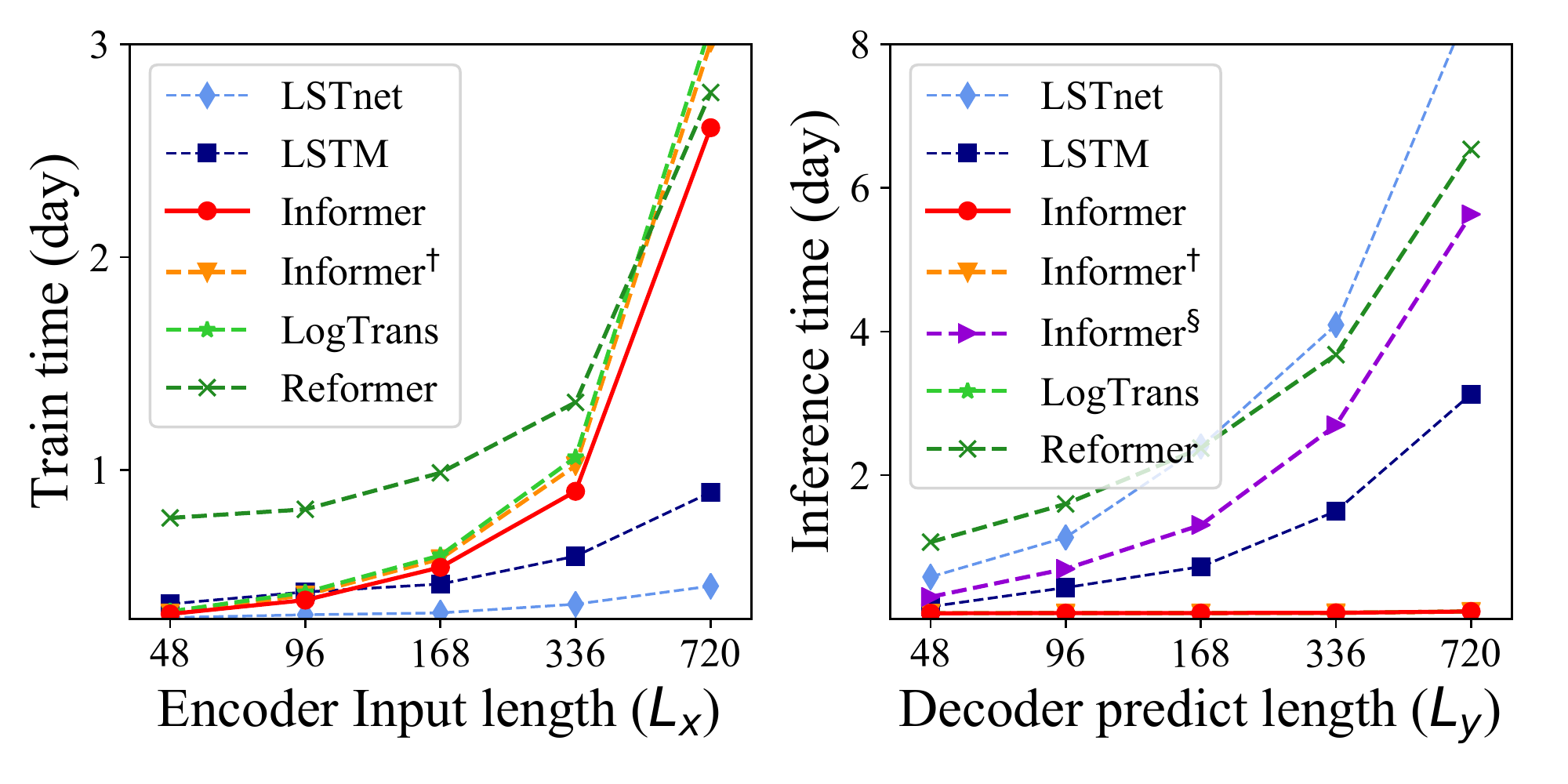}
  \caption{The total runtime of training/testing phase.}
  \label{fig:exp.runtime_new}
  \vspace{-2 ex}
\end{figure}

\textbf{The performance of generative style decoder}
In this study, we testify the potential value of our decoder in acquiring a ``generative'' results. Unlike the existing methods, the labels and outputs are forced to be aligned in the training and inference, our proposed decoder's predicting relies solely on the time stamp, which can predict with offsets. From Table \ref{tab:exp.ablation.decoder}, we can see that the general prediction performance of \mn$^{\ddag}$ resists with the offset increasing, while the counterpart fails for the dynamic decoding. It proves the decoder's ability to capture individual long-range dependency between arbitrary outputs and avoid error accumulation.

\subsection{Computation Efficiency}
\label{sec:Exp.CompuationEfficiency}
With the multivariate setting and all the methods' current finest implement, we perform a rigorous runtime comparison in Fig.(\ref{fig:exp.runtime_new}). During the training phase, the {\mn} (red line) achieves the best training efficiency among Transformer-based methods. During the testing phase, our methods are much faster than others with the generative style decoding. The comparisons of theoretical time complexity and memory usage are summarized in Table \ref{tab:exp.computation.summary}. The performance of {\mn} is aligned with the runtime experiments. Note that the LogTrans focus on improving the self-attention mechanism, and we apply our proposed decoder in LogTrans for a fair comparison (the $\star$ in Table \ref{tab:exp.computation.summary}).

\section{Conclusion}
In this paper, we studied the long-sequence time-series forecasting problem and  proposed {\mn} to predict long sequences. Specifically, we designed the \emph{ProbSparse} self-attention mechanism and distilling operation to handle the challenges of quadratic time complexity and quadratic memory usage in vanilla Transformer. Also, the carefully designed generative decoder alleviates the limitation of traditional encoder-decoder architecture. The experiments on real-world data demonstrated the effectiveness of {\mn} for enhancing the prediction capacity in LSTF problem.

\begin{appendices}

\section{Related Work}
\label{sec:appendix.relatedwork}
We provide a literature review of the long sequence time-series forecasting (LSTF) problem below.

\textbf{Time-series Forecasting} Existing methods for time-series forecasting can be roughly grouped into two categories: classical models and deep learning based methods.
Classical time-series models serve as a reliable workhorse for time-series forecasting, with appealing properties such as interpretability and theoretical guarantees \cite{box2015time,ray1990time}.
Modern extensions include the support for missing data \cite{seeger2017approximate} and multiple data types \cite{seeger2016bayesian}.
Deep learning based methods mainly develop sequence to sequence prediction paradigm by using RNN and their variants, achieving ground-breaking performance \cite{hochreiter1997long,li2017graph,yu2017long}.
Despite the substantial progress, existing algorithms still fail to predict long sequence time series with satisfying accuracy. Typical state-of-the-art approaches \cite{seeger2017approximate,seeger2016bayesian}, especially deep-learning methods \cite{yu2017long,conf/ijcai/QinSCCJC17,flunkert2017deepar,mukherjee2018armdn,wen2017multi}, remain as a sequence to sequence prediction paradigm with step-by-step process, which have the following limitations: (i) Even though they may achieve accurate prediction for one step forward, they often suffer from accumulated error from the dynamic decoding, resulting in the large errors for LSTF problem \cite{journals/corr/abs-1904-07464,conf/ijcai/QinSCCJC17}. The prediction accuracy decays along with the increase of the predicted sequence length. (ii) Due to the problem of vanishing gradient and memory constraint \cite{sutskever2014sequence}, most existing methods cannot learn from the past behavior of the whole history of the time-series. In our work, the {\mn} is designed to address this two limitations.

\textbf{Long sequence input problem} From the above discussion, we refer to the second limitation as to the long sequence time-series input (LSTI) problem. We will explore related works and draw a comparison between our LSTF problem. The researchers truncate / summarize / sample the input sequence to handle a very long sequence in practice, but valuable data may be lost in making accurate predictions. Instead of modifying inputs, Truncated BPTT \cite{aicher2019adaptively} only uses last time steps to estimate the gradients in weight updates, and Auxiliary Losses \cite{trinh2018learning} enhance the gradients flow by adding auxiliary gradients. Other attempts includes Recurrent Highway Networks \cite{zilly2017recurrent} and Bootstrapping Regularizer \cite{cao2019better}. Theses methods try to improve the gradient flows in the recurrent network's long path, but the performance is limited with the sequence length growing in the LSTI problem. CNN-based methods \cite{stoller2019seq,bai2018convolutional} use the convolutional filter to capture the long term dependency, and their receptive fields grow exponentially with the stacking of layers, which hurts the sequence alignment. In the LSTI problem, the main task is to enhance the model's capacity of receiving long sequence inputs and extract the long-range dependency from these inputs. But the LSTF problem seeks to enhance the model's prediction capacity of forecasting long sequence outputs, which requires establishing the long-range dependency between outputs and inputs. Thus, the above methods are not feasible for LSTF directly.

\textbf{Attention model} Bahdanau et al. firstly proposed the addictive attention \cite{bahdanau2014neural} to improve the word alignment of the encoder-decoder architecture in the translation task. Then, its variant \cite{DBLP:conf/emnlp/LuongPM15} has proposed the widely used location, general, and dot-product attention.
The popular self-attention based Transformer \cite{vaswani2017attention} has recently been proposed as new thinking of sequence modeling and has achieved great success, especially in the NLP field. The ability of better sequence alignment has been validated by applying it to translation, speech, music, and image generation. In our work, the {\mn} takes advantage of its sequence alignment ability and makes it amenable to the LSTF problem.

\textbf{Transformer-based time-series model} The most related works \cite{song2018attend,ma2019cdsa,li2019enhancing} all start from a trail on applying Transformer in time-series data and fail in LSTF forecasting as they use the vanilla Transformer. And some other works \cite{child2019generating, li2019enhancing} noticed the sparsity in self-attention mechanism and we have discussed them in the main context.

\section{The Uniform Input Representation}
\label{sec:appendix.input}
\begin{figure}[h]
  \centering
  \includegraphics[width=0.85\linewidth]{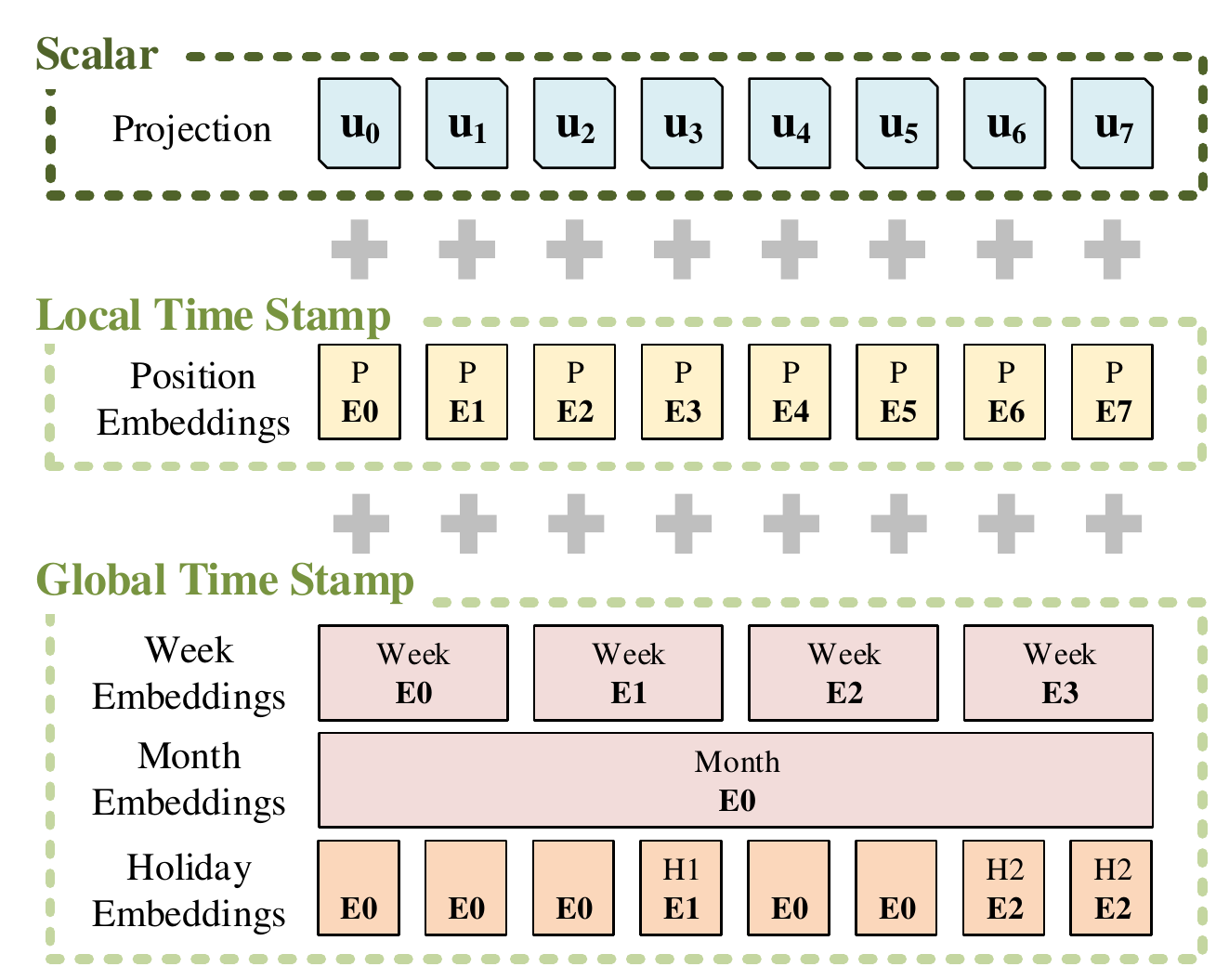}
  \caption{The input representation of \mn. The inputs's embedding consists of three separate parts, a scalar projection, the local time stamp (Position) and global time stamp embeddings (Minutes, Hours, Week, Month, Holiday etc.).}
  \label{fig:method.embed}
\end{figure}

The RNN models \cite{schuster1997bidirectional,hochreiter1997long,chung2014empirical,sutskever2014sequence,conf/ijcai/QinSCCJC17,chang2018memory} capture the time-series pattern by the recurrent structure itself and barely relies on time stamps. The vanilla transformer \cite{vaswani2017attention, devlin2018bert} uses point-wise self-attention mechanism and the time stamps serve as local positional context. However, in the LSTF problem, the ability to capture long-range independence requires global information like hierarchical time stamps (week, month and year) and agnostic time stamps (holidays, events). These are hardly leveraged in canonical self-attention and consequent query-key mismatches between the encoder and decoder bring underlying degradation on the forecasting performance. We propose a uniform input representation to mitigate the issue, the Fig.(\ref{fig:method.embed}) gives an intuitive overview.


Assuming we have $t$-th sequence input $\mathcal{X}^t$ and $p$ types of global time stamps and the feature dimension after input representation is $d_{\rm{model}}$. We firstly preserve the local context by using a fixed position embedding:
\begin{equation}
\begin{aligned}
   \textrm{PE}_{(pos, 2j)} = \sin(pos/(2L_x)^{2j/d_{\rm{model}}}) \\
   \textrm{PE}_{(pos, 2j+1)} = \cos(pos/(2L_x)^{2j/d_{\rm{model}}})
\end{aligned}\quad,
\end{equation}
where $j\in\{1, \dots, \lfloor {d_{\rm{model}}}/{2} \rfloor \}$. Each global time stamp is employed by a learnable stamp embeddings $\textrm{SE}_{(pos)}$ with limited vocab size (up to 60, namely taking minutes as the finest granularity). That is, the self-attention's similarity computation can have access to global context and the computation consuming is affordable on long inputs. To align the dimension, we project the scalar context $\mb{x}^t_{i}$ into $d_{\rm{model}}$-dim vector $\mb{u}^t_{i}$ with 1-D convolutional filters (kernel width=3, stride=1). Thus, we have the feeding vector
\begin{small}
\begin{equation}\label{eq:method.input.sum}
  \mathcal{X}^t_{\textrm{feed}[i]} = \alpha \mb{u}^t_i + \textrm{PE}_{(L_x \times (t-1)+i,~)}  \!+\! \sum_{p} [\textrm{SE}_{(L_x \times (t-1)+i)}]_p \quad,
\end{equation}
\end{small}
where $i\in\{1, \dots, L_x\}$, and $\alpha$ is the factor balancing the magnitude between the scalar projection and local/global embeddings. We recommend $\alpha=1$ if the sequence input has been normalized.

\section{The long tail distribution in self-attention feature map}
\begin{figure}[ht]
  \centering
  \includegraphics[width=0.95\linewidth]{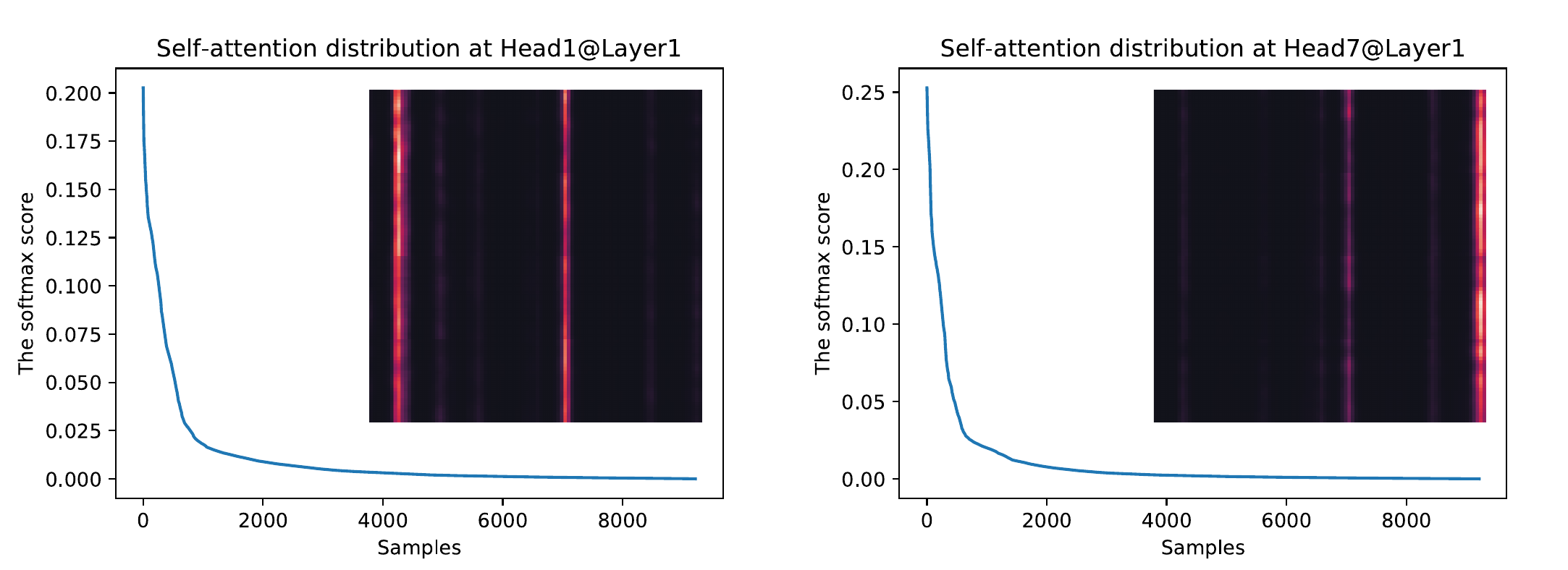}
  \caption{The Softmax scores in the self-attention from a 4-layer canonical Transformer trained on \textbf{ETTh$_1$} dataset.}
  \label{fig:method.self-attention.score}
\end{figure}

We have performed the vanilla Transformer on the \textbf{ETTh$_1$} dataset to investigate the distribution of self-attention feature map. We select the attention score of \{Head1,Head7\} @ Layer1. The blue line in Fig.(\ref{fig:method.self-attention.score}) forms a long tail distribution, i.e. a few dot-product pairs contribute to the major attention and others can be ignored.
\label{sec:appendix.longtail}

\section{Details of the proof}
\subsection{Proof of Lemma 1}
\label{sec:appendix.proof1}
\begin{proof}
For the individual $\mb{q}_i$, we can relax the discrete keys into the continuous $d$-dimensional variable, i.e. vector $\mb{k}_j$. The query sparsity measurement is defined as the $M(\mb{q}_i, \mb{K}) = \ln \sum_{j=1}^{L_K} e^{{\mb{q}_i\mb{k}_j^{\top}}/{\sqrt{d}}} - \frac{1}{L_K} \sum_{j=1}^{L_K} ({\mb{q}_i\mb{k}_j^{\top}}/{\sqrt{d}})$.

Firstly, we look into the left part of the inequality. For each query $\mb{q}_i$,  the first term of the $M(\mb{q}_i, \mb{K})$ becomes the log-sum-exp of the inner-product of a fixed query $\mb{q}_i$ and all the keys , and we can define $f_i(\mb{K})=\ln \sum_{j=1}^{L_K} e^{{\mb{q}_i\mb{k}_j^{\top}}/{\sqrt{d}}}$. From the Eq.(2) in the Log-sum-exp network\cite{DBLP:journals/corr/abs-1806-07850} and the further analysis, the function $f_i(\mb{K})$ is convex. Moreover, $f_i(\mb{K})$ add a linear combination of $\mb{k}_j$ makes the $M(\mb{q}_i, \mb{K})$ to be the convex function for a fixed query. Then we can take the derivation of the measurement with respect to the individual vector $\mb{k}_j$ as $\frac{\partial M(\mb{q}_i, \mb{K})}{\partial \mb{k}_j} = \frac{e^{\mb{q}_i\mb{k}_j^{\top}/{\sqrt{d}}}}{\sum_{j=1}^{L_K} e^{\mb{q}_i\mb{k}_j^{\top}/{\sqrt{d}}}} \cdot \frac{\mb{q}_i}{\sqrt{d}} - \frac{1}{L_K} \cdot \frac{\mb{q}_i}{\sqrt{d}}$.
To reach the minimum value, we let $\vec{\nabla} M(\mb{q}_i) = \vec{0}$ and the following condition is acquired as $\mb{q}_i\mb{k}_1^{\top} + \ln L_K = \cdots = \mb{q}_i\mb{k}_j^{\top} + \ln L_K = \cdots = \ln \sum_{j=1}^{L_K} e^{\mb{q}_i\mb{k}_j^{\top}}$.
Naturally, it requires $\mb{k}_1=\mb{k}_2=\cdots=\mb{k}_{L_K}$, and we have the measurement's minimum as $\ln L_K$, i.e.
\begin{equation}
\label{eq.appendix.proof1.left}
    M(\mb{q}_i, \mb{K}) \geq \ln L_K \qquad.
\end{equation}

Secondly, we look into the right part of the inequality. If we select the largest inner-product $\max_{j}\{{\mb{q}_i \mb{k}_j^{\top}}/{\sqrt{d}}\}$, it is easy that

\begin{small}
\begin{equation}
\begin{aligned}
    M(\mb{q}_i, \mb{K}) &= \ln \sum_{j=1}^{L_K} e^{\frac{{\mb{q}_i\mb{k}_j^{\top}}}{{\sqrt{d}}}} \!-\!\frac{1}{L_K} \sum_{j=1}^{L_K} (\frac{{\mb{q}_i\mb{k}_j^{\top}}}{\sqrt{d}}) \\
    & \leq \ln ( L_K \!\cdot\! \max_{j}\{\frac{{\mb{q}_i \mb{k}_j^{\top}}}{\sqrt{d}}\} )\!-\!\frac{1}{L_K} \sum_{j=1}^{L_K} (\frac{\mb{q}_i\mb{k}_j^{\top}}{\sqrt{d}}) \\
    & = \ln L_K \!+\! \max_{j}\{\frac{\mb{q}_i \mb{k}_j^{\top}}{\sqrt{d}}\}\!-\!
    \frac{1}{L_K} \sum_{j=1}^{L_K} (\frac{\mb{q}_i\mb{k}_j^{\top}}{\sqrt{d}})
\end{aligned}.
\end{equation}
\end{small}

Combine the Eq.(14) and Eq.(15), we have the results of Lemma 1. When the key set is the same with the query set, the above discussion also holds.
\end{proof}

\begin{proposition}
Assuming $\mb{k}_j \sim \mathcal{N} (\mu, \Sigma)$ and we let $\mb{qk}_{i}$ denote set $\{({\mb{q}_i\mb{k}_j^{\top}})/{\sqrt{d}} ~|~ j=1,\ldots,L_K\}$, then $\forall M_{m} = \max_{i} M(\mb{q}_i, \mb{K})$ there exist $\kappa > 0$ such that: in the interval $\forall \mb{q}_1, \mb{q}_2 \in \{ \mb{q} | M(\mb{q}, \mb{K}) \in [ M_{m},  M_{m} - \kappa)\}$, if $\overline{M} (\mb{q}_1, \mb{K}) > \overline{M} (\mb{q}_2, \mb{K})$ and~$\mathrm{Var}(\mb{qk}_{1}) > \mathrm{Var}(\mb{qk}_{2})$, we have high probability that $M(\mb{q}_1, \mb{K}) > M(\mb{q}_2, \mb{K})$.
\end{proposition}

\subsection{Proof of Proposition 1}
\label{sec:appendix.proof2}
\begin{proof}
To make the further discussion simplify, we can note $a_{i,j}={q_i k_j^T}/{\sqrt{d}}$, thus define the array $A_i = [a_{i,1},\cdots,a_{i,L_k}]$. Moreover, we denote $\frac{1}{L_{K}} \sum_{j=1}^{L_{K}} ({\mathbf{q}_{i} \mathbf{k}_{j}^{\top}}/{\sqrt{d}})=\mathrm{mean}(A_i)$, then we can denote $\bar{M}\left(\mathbf{q}_{i}, \mathbf{K}\right)=\max(A_i)-\mathrm{mean}(A_i)$, $i=1,2$.

As for $M\left(\mathbf{q}_{i}, \mathbf{K}\right)$, we denote each component $a_{i,j}= \mathrm{mean}(A_i) + \Delta a_{i,j},j=1,\cdots,L_k$, then we have the following:
\begin{equation}
\nonumber
\begin{aligned}
M\left(\mathbf{q}_{i}, \mathbf{K}\right)=&\ln \sum_{j=1}^{L_{K}} e^{{\mathbf{q}_{i} \mathbf{k}_{j}^{\top}}/{\sqrt{d}}}-\frac{1}{L_{K}}
\sum_{j=1}^{L_{K}} ({\mathbf{q}_{i} \mathbf{k}_{j}^{\top}}/{\sqrt{d}})\\
=&\ln(\Sigma_{j=1}^{L_k} e^{\mathrm{mean}(A_i)}e^{\Delta a_{i,j}})-\mathrm{mean}(A_i)\\
=&\ln(e^{\mathrm{mean}(A_i)} \Sigma_{j=1}^{L_k} e^{\Delta a_{i,j}})-\mathrm{mean}(A_i)\\
=&\ln(\Sigma_{j=1}^{L_k}e^{\Delta a_{i,j}})
\end{aligned},
\end{equation}
and it is easy to find $\Sigma_{j=1}^{L_k}\Delta a_{i,j} = 0$.

We define the function $ES(A_i)=\Sigma_{j=1}^{L_k}\exp(\Delta a_{i,j})$, equivalently defines $A_i = [\Delta a_{i,1},\cdots,\Delta a_{i,L_k}]$, and immediately our proposition can be written as the equivalent form:

For $\forall A_1,A_2$, if
\begin{enumerate}[nosep,  leftmargin=0.85cm]
\item $\max(A_1)-\mathrm{mean}(A_1)\geq \max(A_2)-\mathrm{mean}(A_2)$
\item $\mathrm{Var}(A_1) > \mathrm{Var}(A_2)$
\end{enumerate}
Then we rephrase the original conclusion into more general form that $ES(A_1)>ES(A_2)$ with high probability, and the probability have positive correlation with $\mathrm{Var}(A_1)-\mathrm{Var}(A_2)$.\\

Furthermore, we consider a fine case, $\forall M_{m} = \max_{i} M(\mb{q}_i, \mb{K})$ there exist $\kappa > 0$ such that in that interval $\forall \mb{q}_i, \mb{q}_j \in \{ \mb{q} | M(\mb{q}, \mb{K}) \in [ M_{m},  M_{m} - \kappa)\}$ if $\max(A_1)-\mathrm{mean}(A_1)\geq \max(A_2)-\mathrm{mean}(A_2)$ and~$\mathrm{Var}(A_1) > \mathrm{Var}(A_2)$, we have high probability that $M(\mb{q}_1, \mb{K}) > M(\mb{q}_2, \mb{K})$,which is equivalent to $ES(A_1) > ES(A_2)$.

In the original proposition, $\mathbf{k}_{\mathbf{j}} \sim \mathcal{N}(\mu, \Sigma)$ follows multivariate Gaussian distribution, which means that $k_1,\cdots,k_n$ are I.I.D Gaussian distribution, thus defined by the Wiener-khinchin law of large Numbers, $a_{i,j}={q_i k_j^T}/{\sqrt{d}}$ is one-dimension Gaussian distribution with the expectation of 0 if $n\rightarrow \infty$. So back to our definition, $\Delta a_{1,m} \sim N(0,\sigma_1^2),\Delta a_{2,m} \sim N(0,\sigma_2^2),\forall m\in 1,\cdots,L_k$, and our proposition is equivalent to a lognormal-distribution sum problem.
 
A lognormal-distribution sum problem is equivalent to approximating the distribution of $ES(A_1)$ accurately, whose history is well-introduced in the articles \cite{dufresne2008sums},\cite{vargasguzman2005change}. Approximating lognormality of sums of lognormals is a well-known rule of thumb, and no general PDF function can be given for the sums of lognormals. However, \cite{romeo2003broad} and \cite{hcine2015approximation} pointed out that in most cases, sums of lognormals is still a lognormal distribution, and by applying central limits theorem in \cite{beaulieu2011extended}, we can have a good approximation that $ES(A_1)$ is a lognormal distribution, and we have $E(ES(A_1))=ne^{\frac{\sigma_1^2}{2}}$, $\mathrm{Var}(ES(A_1))=ne^{\sigma_1^2}(e^{\sigma_1^2}-1)$. Equally, $E(ES(A_2))=ne^{\frac{\sigma_2^2}{2}}$, $\mathrm{Var}(ES(A_2))=ne^{\sigma_2^2}(e^{\sigma_2^2}-1)$.

We denote $B_1=ES(A_1),B_2=ES(A_2)$, and the probability $Pr(B_1-B_2>0)$ is the final result of our proposition in general conditions, with $\sigma_1^2>\sigma_2^2$ WLOG. The difference of lognormals is still a hard problem to solve.

By using the theorem given in\cite{lo2012sum}, which gives a general approximation of the probability distribution on the sums and difference for the lognormal distribution. Namely $S_1$ and $S_2$ are two lognormal stochastic variables obeying the stochastic differential equations$\frac{d S_{i}}{S_{i}}=\sigma_{i} d Z_{i}$, $i=1,2,$ in which $dZ_{1,2}$ presents a standard Weiner process associated with $S_{1,2}$ respectively, and $\sigma_{i}^{2}=\operatorname{Var}\left(\ln S_{i}\right)$, $S^{\pm} \equiv S_{1} \pm S_{2}$,$S_0^{\pm} \equiv S_{10} \pm S_{20}$. As for the joint probability distribution function $P\left(S_{1}, S_{2}, t ; S_{10}, S_{20}, t_{0}\right)$, the value of $S_1$ and $S_2$ at time $t>t_0$ are provided by their initial value $S_{10}$ and $S_{20}$ at initial time $t_0$. The Weiner process above is equivalent to the lognormal distribution\cite{weiner1984meaning}, and the conclusion below is written in general form containing both the sum and difference of lognormal distribution approximation denoting $\pm$ for sum $+$ and difference $-$ respectively.
 
In boundary condition 
 \begin{equation*}
     \bar{P}_{\pm}\left(S^{\pm}, t ; S_{10}, S_{20}, t_{0} \longrightarrow t\right)=\delta\left(S_{10} \pm S_{20}-S^{\pm}\right) \quad,
 \end{equation*}
their closed-form probability distribution functions are given by 
  \begin{equation*}
  \begin{aligned}
     &\qquad f^{\mathrm{LN}}\left(\tilde{S}^{\pm}, t ; \tilde{S}_{0}^{\pm}, t_{0}\right)\\
     = &\frac{1}{\widetilde{S}^{\pm} \sqrt{2 \pi \tilde{\sigma}_{\pm}^{2}\left(t-t_{0}\right)}} \\
      &\cdot \exp \left\{-\frac{\left[\ln \left(\tilde{S}^{+} / \tilde{S}_{0}^{+}\right)+(1 / 2) \tilde{\sigma}_{\pm}^{2}\left(t-t_{0}\right)\right]^{2}}{2 \tilde{\sigma}_{\pm}^{2}\left(t-t_{0}\right)}\right\} \quad.
 \end{aligned}
 \end{equation*}
It is an approximately normal distribution, and $\tilde{S}^{+}$, $\tilde{S}^{-}$ are lognormal random variables, $\tilde{S}_{0}^{\pm}$ are initial condition in $t_0$ defined by Weiner process above. (Noticed that $\tilde{\sigma}_{\pm}^{2}\left(t-t_{0}\right)$ should be small to make this approximation valid.In our simulation experiment, we set $t-t_0=1$ WLOG.) Since
\begin{equation*}
\widetilde{S}_{0}^{-}=(S_{10}-S_{20})+\left(\frac{\sigma_{-}^{2}}{\sigma_{1}^{2}-\sigma_{2}^{2}}\right) (S_{10}+S_{20}),\\ 
\end{equation*}
and
\begin{equation*}
    \tilde{\sigma_-}=\left(\sigma_{1}^{2}-\sigma_{2}^{2}\right) /\left(2 \sigma_{-}\right)\\
\end{equation*}
\begin{equation*}
    \sigma_-=\sqrt{\sigma_1^2+\sigma_2^2}
\end{equation*}
Noticed that $E(B_1)>E(B_2)$, $\mathrm{Var}(B_1)>\mathrm{Var}(B_2)$, the mean value and the variance of the approximate normal distribution shows positive correlation with $\sigma_1^2-\sigma_2^2$.Besides, the closed-form PDF $f^{\mathrm{LN}}\left(\tilde{S}^{\pm}, t ; \tilde{S}_{0}^{\pm}, t_{0}\right)$ also show positive correlation with $\sigma_1^2-\sigma_2^2$. Due to the limitation of $\tilde{\sigma}_{\pm}^{2}\left(t-t_{0}\right)$ should be small enough, such positive correlation is not significant in our illustrative numerical experiment.

By using Lie-Trotter Operator Splitting Method in \cite{lo2012sum}, we can give illustrative numeral examples for the distribution of $B_1-B_2$,in which the parameters are well chosen to fit for our top-u approximation in actual LLLT experiments. Figure shows that it is of high probability that when $\sigma_1^2>\sigma_2^2$, the inequality holds that $B_1>B_2$, $ES(A_1)>ES(A_2)$.

Finishing prooving our proposition in general conditions, we can consider a more specific condition that if $\mb{q}_1, \mb{q}_2 \in \{ \mb{q} | M(\mb{q}, \mb{K}) \in [ M_{m},  M_{m} - \kappa)\}$, the proposition still holds with high probability.

First, we have $M(q_1,\mb{k})=ln(B_1)>(M_m-\kappa)$ holds for $\forall q_1,q_2$ in this interval. Since we have proved that $E(B_1))=ne^{\frac{\sigma_1^2}{2}}$, we can conclude that $\forall q_i$ in the given interval,$\exists \alpha, \sigma_i^2>\alpha, i=1,2$. Since we have $\widetilde{S}_{0}^{-}=(S_{10}-S_{20})+\left(\frac{\sigma_{-}^{2}}{\sigma_{1}^{2}-\sigma_{2}^{2}}\right) (S_{10}+S_{20})$, which also shows positive correlation with $\sigma_1^2+\sigma_2^2>2\alpha$, and positive correlation with $\sigma_1^2-\sigma_2^2$. So due to the nature of the approximate normal distribution PDF, if $\sigma_1^2>\sigma_2^2$ WLOG, $Pr(M(q_1,\mb{k})>M(q_2,\mb{k}))\approx \Phi(\frac{\widetilde{S}_{0}^{-}}{\tilde{\sigma_-}})$ also shows positive correlation with $\sigma_1^2+\sigma_2^2>2\alpha$.

\begin{figure}
    \resizebox{0.85\linewidth}{!}{\includegraphics{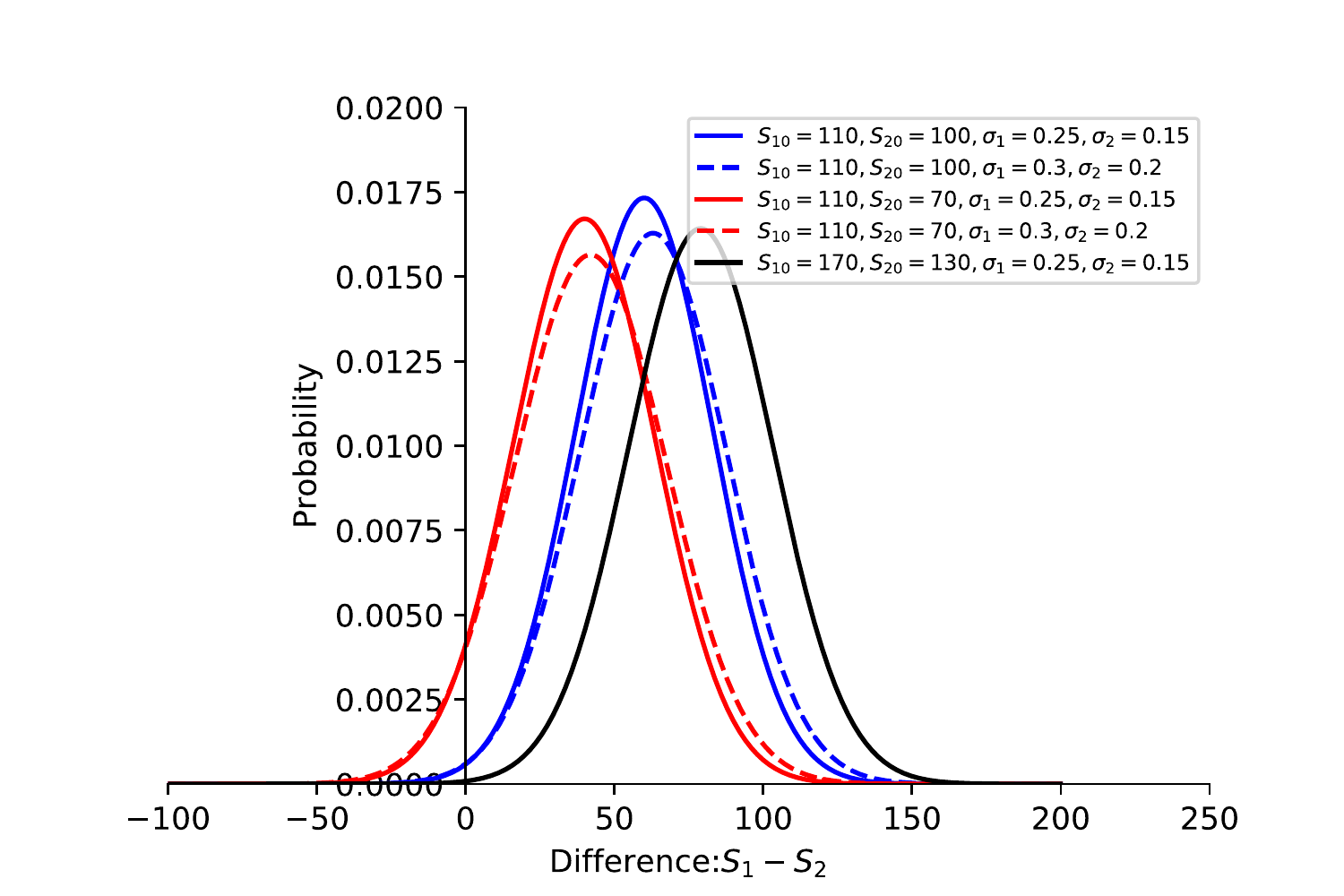}}
    \centering
    \caption{Probability Density verses $S_1-S_2$ for the approximation of  shifted lognormal distribution.}
    \label{fig:appendix.examples}
\end{figure}

We give an illustrative numerical examples of the approximation above in Fig.(\ref{fig:appendix.examples}). In our actual LTTnet experiment, we choose Top-k of $A_1,A_2$, not the whole set.Actually, we can make a naive assumption that in choosing $top-\lfloor\frac{1}{4}L_k\rfloor$ variables of $A_1,A_2$ denoted as $A_1^{'} , A_2^{'}$,the variation $\sigma_1,\sigma_2$ don't change significantly, but the expectation $E(A_1^{'}) , E(A_2^{'})$ 
ascends obviously, which leads to initial condition $S_{10},S_{20}$ ascends significantly, since the initial condition will be sampled from $top-\lfloor\frac{1}{4}L_k\rfloor$ variables, not the whole set.

In our actual LTTnet experiment, we set $U$, namely choosing around $top-\lfloor\frac{1}{4}L_k\rfloor$ of $A_1$ and $A_2$, it is guaranteed that with over $99\%$ probability that in the $[ M_{m},  M_{m} - \kappa)$ interval, as shown in the black curve of Fig.(\ref{fig:appendix.examples}). Typically the condition 2 can be relaxed, and we can believe that if $q_1,q_2$ fits the condition 1 in our proposition, we have $M(\mb{q}_1, \mb{K}) > M(\mb{q}_2, \mb{K})$.
\end{proof}

\section{Reproducibility}
\subsection{Details of the experiments}
\label{sec:appendix.network}
The details of proposed \mn~ model is summarized in Table \ref{tab:method.network.detail}. For the \emph{ProbSparse} self-attention mechanism, we let $d$=32, $n$=16 and add residual connections, a position-wise feed-forward network layer (inner-layer dimension is 2048) and a dropout layer ($p=0.1$) likewise. Note that we preserves 10\% validation data for each dataset, so all the experiments are conducted over 5 random train/val shifting selection along time and the results are averaged over the 5 runs. All the datasets are performed standardization such that the mean of variable is 0 and the standard deviation is 1.

\begin{table}[ht]
\caption{The \mn~ network components in details}
\label{tab:method.network.detail}
\small
\centering
\begin{tabular}{c|c|c|c}
\hline
\hline
\multicolumn{3}{l}{\textbf{Encoder:}} & N \\
\hline
Inputs & 1x3 Conv1d & Embedding ($d=512$) & \multirow{7}{*}{4} \\
\cline{1-3}
\multirow{4}{*}{\shortstack{ProbSparse\\ Self-attention\\ Block }} & \multicolumn{2}{|c|}{\scriptsize{Multi-head ProbSparse Attention} ($h=16$, $d=32$)} & \\
\cline{2-3}
 & \multicolumn{2}{|c|}{Add, LayerNorm, Dropout ($p=0.1$)} & \\
\cline{2-3}
 & \multicolumn{2}{|c|}{Pos-wise FFN ($d_{\rm{inner}}=2048$), GELU } & \\
\cline{2-3}
 & \multicolumn{2}{|c|}{Add, LayerNorm, Dropout ($p=0.1$)} & \\
\cline{1-3}
\multirow{2}{*}{Distilling} & \multicolumn{2}{|c|}{1x3 conv1d, ELU} & \\
\cline{2-3}
 & \multicolumn{2}{|c|}{Max pooling ($\textrm{stride}=2$)} & \\
\hline
\multicolumn{3}{l}{\textbf{Decoder:}} & N \\
\hline
Inputs & 1x3 Conv1d & Embedding ($d=512$) & \multirow{6}{*}{2} \\
\cline{1-3}
Masked PSB & \multicolumn{2}{|c|}{add Mask on Attention Block} & \\
\cline{1-3}
\multirow{4}{*}{\shortstack{Self-attention\\ \\ Block }} & \multicolumn{2}{|c|}{Multi-head Attention ($h=8$, $d=64$)} & \\
\cline{2-3}
 & \multicolumn{2}{|c|}{Add, LayerNorm, Dropout ($p=0.1$)} & \\
\cline{2-3}
 & \multicolumn{2}{|c|}{Pos-wise FFN ($d_{\rm{inner}}=2048$), GELU } & \\
\cline{2-3}
 & \multicolumn{2}{|c|}{Add, LayerNorm, Dropout ($p=0.1$)} & \\
\hline
\multicolumn{4}{l}{\textbf{Final:}} \\
\hline
Outputs & \multicolumn{2}{|c|}{FCN ($d=d_{\rm{out}}$)} & \\
\cline{1-3}
\hline
\end{tabular}
\end{table}

\subsection{Implement of the \emph{ProbSparse} self-attention}
\label{sec:appendix.algo}
We have implemented the \emph{ProbSparse} self-attention in Python 3.6 with Pytorch 1.0. The pseudo-code is given in Algo.(\ref{alg:mPegasos.pbAttention}). The source code is available at \url{https://github.com/zhouhaoyi/Informer2020}. All the procedure can be highly efficient vector operation and maintains logarithmic total memory usage. The masked version can be achieved by applying positional mask on step 6 and using $\textrm{cmusum}(\cdot)$ in $\textrm{mean}(\cdot)$ of step 7. In the practice, we can use $\textrm{sum}(\cdot)$ as the simpler implement of $\textrm{mean}(\cdot)$.

\begin{algorithm}[ht]
\caption{ProbSparse self-attention}
\label{alg:mPegasos.pbAttention}
\begin{algorithmic}[1]
\small
\REQUIRE
Tensor $\mb{Q}\in \mathbb{R}^{m \times d}$, $\mb{K}\in \mathbb{R}^{n \times d}$, $\mb{V}\in \mathbb{R}^{n \times d}$\\
\PRINT set hyperparameter $c$, $u= c \ln m$ and $U= m \ln n$
\STATE randomly select $U$ dot-product pairs from $\mb{K}$ as $\bar{\mb{K}}$
\STATE set the sample score $\bar{\mathbf{S}} = \mb{Q} \bar{\mb{K}}^{\top}$
\STATE compute the measurement $M= \textrm{max} (\bar{\mathbf{S}}) - \textrm{mean} (\bar{\mathbf{S}})$ by row
\STATE set Top-$u$ queries under $M$ as $\bar{\mb{Q}}$
\STATE set $\mathbf{S}_1 = \textrm{softmax}(\bar{\mb{Q}} \mb{K}^{\top} / \sqrt{d}) \cdot \mb{V}$
\STATE set $\mathbf{S}_0 = \textrm{mean}(\mb{V})$
\STATE set $\mathbf{S} = \{\mathbf{S}_1,\mathbf{S}_0\}$ by their original rows accordingly
\ENSURE
self-attention feature map $\mathbf{S}$. \\
\end{algorithmic}
\end{algorithm}

\begin{figure*}[t]
  \centering
  \includegraphics[width=1\linewidth]{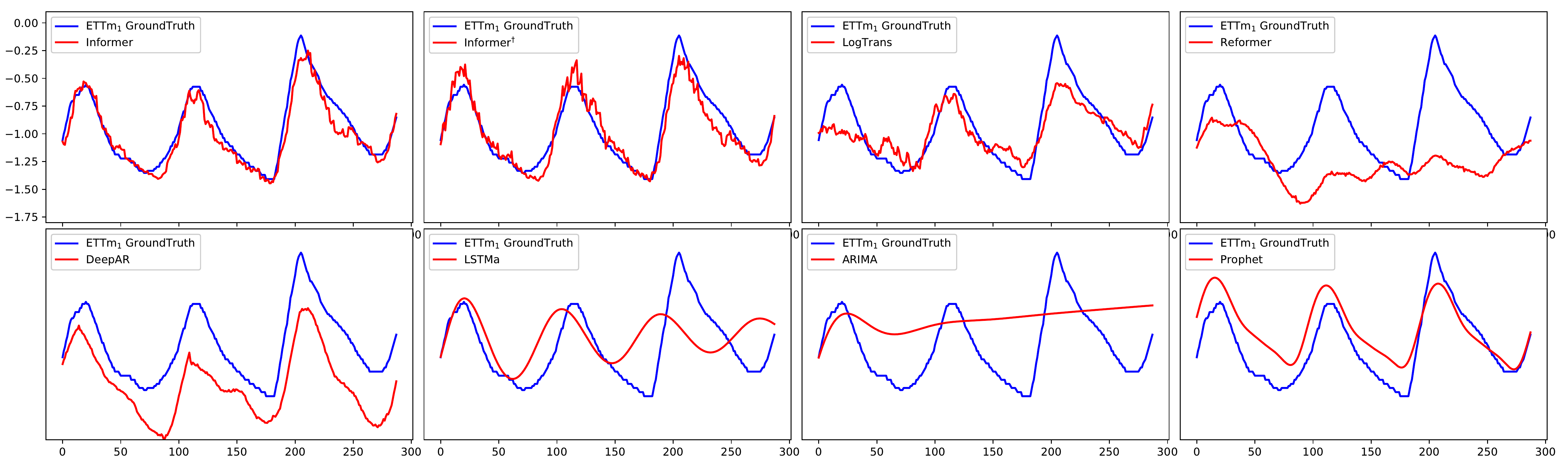}
  \caption{The predicts (len=336) of \mn, \mn$^{\dag}$, LogTrans, Reformer, DeepAR, LSTMa, ARIMA and Prophet on the ETTm dataset. The red / blue curves stand for slices of the prediction / ground truth.}
  \label{fig:appendix.exp.curve}
\end{figure*}

\subsection{The hyperparameter tuning range}
\label{sec:appendix.range}
For all methods, the input length of recurrent component is chosen from \{24, 48, 96, 168, 336, 720\} for the ETTh1, ETTh2, Weather and Electricity dataset, and chosen from \{24, 48, 96, 192, 288, 672\} for the ETTm dataset. 
For LSTMa and DeepAR, the size of hidden states is chosen from \{32, 64, 128, 256\}. 
For LSTnet, the hidden dimension of the Recurrent layer and Convolutional layer is chosen from \{64, 128, 256\} and \{32, 64, 128\} for Recurrent-skip layer, and the skip-length of Recurrent-skip layer is set as 24 for the ETTh1, ETTh2, Weather and ECL dataset, and set as 96 for the ETTm dataset.
For {\mn}, the layer of encoder is chosen from \{6, 4, 3, 2\} and the layer of decoder is set as 2. The head number of multi-head attention is chosen from \{8, 16\}, and the dimension of multi-head attention’s output is set as 512. The length of encoder’s input sequence and decoder’s start token is chosen from \{24, 48, 96, 168, 336, 480, 720\} for the ETTh1, ETTh2, Weather and ECL dataset, and \{24, 48, 96, 192, 288, 480, 672\} for the ETTm dataset. In the experiment, the decoder’s start token is a segment truncated from the encoder’s input sequence, so the length of decoder’s start token must be less than the length of encoder’s input.

The RNN-based methods perform a dynamic decoding with left shifting on the prediction windows. Our proposed methods \mn-series and LogTrans (our decoder) perform non-dynamic decoding.

\section{Extra experimental results}

Fig.(\ref{fig:appendix.exp.curve}) presents a slice of the predicts of 8 models. The most realted work LogTrans and Reformer shows acceptable results. The LSTMa model is not amenable for the long sequence prediction task. The ARIMA and DeepAR can capture the long trend of the long sequences. And the Prophet detects the changing point and fits it with a smooth curve better than the ARIMA and DeepAR. Our proposed model \mn~ and \mn$^{\dag}$ show significantly better results than above methods.

\section{Computing Infrastructure}
All the experiments are conducted on Nvidia Tesla V100 SXM2 GPUs (32GB memory). Other configuration includes 2 * Intel Xeon Gold 6148 CPU, 384GB DDR4 RAM and 2 * 240GB M.2 SSD, which is sufficient for all the baselines. 

\end{appendices}

\bibliographystyleapndx{aaai21}
\bibliography{LaTeX_arxiv}

\end{document}